\documentclass[journal]{IEEEtran}
\usepackage{amsmath,amsfonts}
\usepackage{lipsum}
\usepackage{balance}
\usepackage{algorithmic}
\usepackage{algorithm}
\usepackage{array}
\usepackage[caption=false,font=normalsize,labelfont=sf,textfont=sf]{subfig}
\usepackage{graphicx}
\usepackage{textcomp}
\usepackage{stfloats}
\usepackage{url}
\usepackage{verbatim}
\usepackage{graphicx}
\usepackage{cite}
\usepackage{tikz}
\usepackage{enumitem}

\newtheorem{theorem}{Theorem}
\newtheorem{definition}{Definition}

\newtheorem{remark}{Remark}

\newtheorem{proof1}{Proof of Theorem}
\newtheorem{corollary}{Corollary}
\newtheorem{lemma}{Lemma}
\newtheorem{proof2}{}
\usepackage{soul} 
\usepackage{siunitx}
\usepackage{xcolor,soul,framed} 


\newcommand\copyrighttext{%
	\scalebox{0.95}{%
		\parbox{\dimexpr\textwidth-\fboxsep-\fboxrule\relax}{%
			\scriptsize
			\setlength{\baselineskip}{0.9\baselineskip}
			\textcopyright \the\year{} IEEE. Personal use of this material is permitted. Permission from IEEE must be obtained for all other uses, in any current or future media, including reprinting/republishing this material for advertising or promotional purposes, creating new collective works, for resale or redistribution to servers or lists, or reuse of any copyrighted component of this work in other works.\\
			This is the author's accepted version of the article: V. T. Dang, H. Kotake, S. Honji, and T. Wada, "A Cooperation Control Framework Based on Admittance Control and Time-varying Passive Velocity Field Control for Human–Robot Co-carrying Tasks," \textit{IEEE Transactions on Automation Science and Engineering}, 2025. DOI:10.1109/TASE.2025.3628318.%
		}%
	}%
}

\newcommand\copyrightnotice{%
	\begin{tikzpicture}[remember picture, overlay]
	\node[anchor=north, yshift=-4pt] at (current page.north)
	{\textcolor{red}{\copyrighttext}};
	\end{tikzpicture}%
}

\begin{document}

\title{A Cooperation Control Framework Based on Admittance Control and Time-varying Passive Velocity Field Control for Human--Robot Co-carrying Tasks}

\author{

\author{Van Trong Dang,~\IEEEmembership{Student Member,~IEEE}, Hiroki Kotake, Sumitaka Honji,~\IEEEmembership{Member,~IEEE}, and Takahiro Wada,~\IEEEmembership{Member,~IEEE}
\thanks{The authors are with the Graduate School of Science and Technology, Nara Institute of Science and Technology, Ikoma, Nara, 630-0192, Japan (e-mail: van\_trong.dang.ve6@naist.ac.jp, kotake.hiroki.kf8@is.naist.jp, honji.sumitaka@naist.ac.jp, and t.wada@is.naist.jp).}
\thanks{}}
\thanks{
} }


\maketitle

\copyrightnotice

\begin{abstract}
Human--robot co-carrying tasks reveal their potential in both industrial and everyday applications by leveraging the strengths of both parties. Effective control of robots in these tasks requires managing the energy level in the closed-loop systems to prevent potential dangers while also minimizing motion errors to complete the shared tasks. The collaborative tasks pose numerous challenges due to varied human intentions in adapting to workspace characteristics, leading to human--robot conflicts. In this paper, we develop a cooperation control framework for human--robot co-carrying tasks constructed by utilizing reference generator and low-level controller to aim to achieve safe interaction and synchronized human--robot movement. Firstly, the human motion predictions are corrected in the event of prediction errors based on the conflicts measured by the interaction forces through admittance control, thereby mitigating conflict levels. Low-level controller using an energy-compensation passive velocity field control approach allows encoding the corrected motion to produce control torques for the robot. In this manner, the closed-loop robotic system is passive when the energy level exceeds the predetermined threshold, and otherwise. Furthermore, the proposed control approach ensures that the system's kinetic energy is compensated within a finite time interval. The passivity, stability, convergence rate of energy, and power flow regulation are analyzed from theoretical viewpoints. Human-in-the-loop experiments involving 18 participants have demonstrated that the proposed method significantly enhances task performance and reduces human workload, as evidenced by both objective metrics and subjective evaluations, with improvements confirmed by statistical tests ($p < 0.05$) relative to baseline methods.
\end{abstract}
\renewcommand{\abstractname}{Note to Practitioners}
\begin{abstract}
This paper is motivated by the challenge of developing a cooperation control framework for human--robot co-carrying tasks in achieving safe interaction and completing the shared tasks, especially in scenarios where human intentions vary due to changes in the environment or task demands. To this end, a reference motion generator, using motion prediction model and admittance model, is first applied to provide human motion intention for the robot in real-time. Subsequently, an energy-compensation passive velocity field control approach is proposed, which utilizes the output of the motion generator to regulate robot behaviors during physical interaction with the human. In this manner, the proposed approach enables the regulation of both the energy level and power flow within the collaborative system, while guiding the robot motion to converge toward the human intention, thereby ensuring safe and synchronized human–robot movement. Through theoretical analysis, the proposed method provides well-defined conditions of control parameters, allowing practitioners to intuitively fine-tune the robot system for specific engineering applications. Furthermore, the proposed method can be extended to other human–robot collaboration tasks, such as object handover, collaborative assembly, and co-sawing, owing to their common features and requirements in physical interaction with the human.
\end{abstract}



\section{Introduction}
\IEEEPARstart{T}{he} field of robotics has recently witnessed numerous studies on physical human--robot interaction (pHRI) problems in the same workspace, such as handover, co-transportation, collaborative assembly, teleoperation, co-sawing, and so on \cite{c1,c2,c3,c3.1,c4}. This arises from the fact that hybrid human--robot systems are considered a potential solution as the cognitive strengths and adaptability of humans can be integrated with the high-precision and computational capability of robots to implement mutual tasks.  However, cooperation failure and increased human workload are common issues in the above human--robot collaboration scenarios \cite{c6.1,c7}, arising from the conflicts between robot behaviors and human intentions. In human--robot collaboration scenarios that involve physical interaction, such as co-carrying tasks, a single object is transported to a target location identified by the human utilizing the mutual efforts of both the robot and human that would be otherwise too large or heavy for a single human to handle \cite{c7.2,c7.1,c7.3}. For such tasks, a leader-follower control framework is often adopted, where the human serves as the leader, and the robot acts as the follower. It is important to achieve safe interaction and synchronize human--robot movement, especially when conflicts arise.

Ensuring safety in pHRI is a critical requirement. A common control approach in such scenarios is to have the robot passively follow the human’s movements when manipulating an object. This is typically achieved through the impedance control or admittance control families \cite{c7.4,c7.111}, which regulate the robot’s response to external forces exerted by the human. Notably, variable impedance control \cite{c7.5,c7.11}, variable admittance control \cite{c7.12}, and fixed admittance control \cite{ c7.15} without a spring term and human motion intention exhibit passive behavior, meaning that the robot does not actively inject energy into the human. However, a drawback of these control approaches is that the human must compensate for closed-loop dynamics of the robot through the object being transported. As a result, the human is required to move both the object and the closed-loop dynamics of the robot, which may not provide meaningful physical assistance in terms of reducing the required effort. To enhance the capability to provide assistance, one possible approach is to integrate the predicted motion into the reference input motion of the impedance control \cite{ce.1,ce.6, c7.8} and admittance control \cite{ce.4,ce.7,c7.6}. In this manner, the robot could proactively assist in the pHRI tasks when the human motion was accurately predicted. However, a major challenge with this method is that it can increase conflict levels and does not inherently guarantee passivity if the accuracy of human intention estimation is low. As a result, the robot may inject energy into the human, potentially compromising safety. Thus, the aforementioned approaches raise fundamental challenges in balancing proactive assistance with guaranteed safety.

In the human--robot co-carrying tasks, an ideal control framework for the robot would satisfy two key properties. To enhance safe interaction, the robot maintains passive behavior in response to external forces exerted by the human when conflict occurs. Meanwhile, the robot is proactive in tracking the target motion trajectory in the absence of external forces for reducing human efforts. Switching control strategies \cite{c7,c7.13} were introduced as possible solutions to deal with the above issues, where the control scheme alternated between passive and active modes based on detected conflicts between human intention and robot movement. Specifically, when conflicts arise, a passive control strategy is employed, whereas an active control strategy is used when no conflicts are detected. However, the switching control methods cause stability concerns due to discontinuities in system behavior.

To address the discontinuous behavior while satisfying the aforementioned properties for human--robot co-carrying tasks, passive velocity field control (PVFC) in \cite{c18,c18.11,c19,c20} has been proposed. Specifically, by using a fictitious flywheel as an energy storage component, the PVFC in \cite{c18,c18.11} fulfills the passive property and global exponential stability of the augmented system. Nevertheless, a drawback of such PVFC is their inherent conservative nature when enforcing tight passivity conditions with respect to external forces, meaning that in scenarios with frequent conflicts, the robot may come to a complete stop, hindering task completion. As a result, the PVFC with an energy compensation control term \cite{c19,c20} was designed to regulate kinetic energy in therapeutic applications. The PVFC in \cite{c18,c18.11,c19,c20} showed remarkable advantages in contour following issues and ensuring safe operation by maintaining the passive relationship between the robot arm under the closed-loop control system and physical environments. However, the existing PVFCs cannot be applied directly in human--robot co-carrying tasks due to the variability of human intention in real-world applications rather than using the predetermined trajectories as reported in \cite{c18,c18.11,c19,c20}. Furthermore, it is necessary to control the energy level in the closed-loop system to guarantee co-carrying performance. This implies that the co-carrying task cannot be achieved at a low energy level due to a sluggish rate, while safety concerns cannot be ensured in the presence of too high energy level.


To address the above issues, the present research proposes a cooperation control framework based on predicted human motion, admittance control, and time-varying energy-compensation PVFC for human--robot co-carrying tasks to enhance task performance, safety interaction, and reduce human workload. The main contributions of the proposed framework are listed as follows:
\begin{enumerate}
\item {Developing a continuous control framework for co-carrying tasks, where the predicted human motion is corrected based on external forces (i.e., conflicts) through admittance control, and the corrected motion is tracked by a time-varying energy-compensating PVFC, it is possible to achieve proactive robotic behavior without the need for discrete switching between control modes \cite{c7,c7.13}.}
\item{The proposed energy-compensation PVFC resolves the conservative nature of conventional PVFC \cite{c18,c18.11} by guaranteeing the passivity of the closed-loop robotic system if and only if the energy level exceeds the predetermined threshold; otherwise, internal energy generation is induced to execute the collaborative tasks.} 
\item {The proposed time-varying PVFC enables not only compensating energy for the closed-loop robotic system within a finite-time interval, but also ensuring precise tracking of the reference motion theoretically in the absence of external forces. Furthermore, the power flow between the robotic system and human can be controlled.}
\end{enumerate}
The rest of the paper is structured in four sections. The problem of human--robot co-carrying tasks and preliminary theories are introduced in the second section. The main innovations are presented in Section III, including the cooperation control framework and its results. In Section IV, the effectiveness of the proposed method is validated through human-in-the-loop experiments. In Section V, the results of the proposed method and its application scope are discussed. Finally, the conclusions and future work are presented in the last section.

\section{Problem Definition and Preliminaries}
\subsection{Problem Definition}

\begin{figure}[!t]
\centering
\includegraphics[scale=0.7]{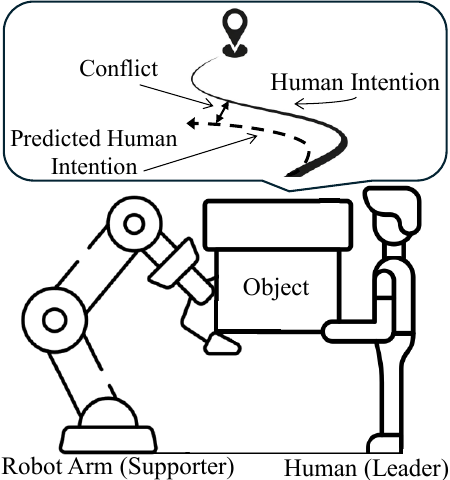}
\caption{Human--robot co-carrying task, in which predicted human intention for the robot may occasionally contain errors or perturbations, thereby causing human--robot conflicts.}
\label{f1}
\end{figure}

In this paper, we consider a human--robot co-carrying task, which is depicted in Fig. \ref{f1}. Specifically, the task involves a collaborative effort where a human and a robot work together to lift and transport a single object. The human acts as a leader of the task thanks to cognitive ability in the workspace, enabling them to devise appropriate strategies and provide guidance to the robot partner. Meanwhile, the robot serves as a supporter, responding to the human's guidance and adjusting its behavior accordingly to assist the human in operating collaborative tasks successfully. For this collaboration scenario, the estimated/ predicted robot's reference motions toward human motion intentions may occasionally contain errors or perturbations due to the variability in human intentions under unstructured environments and the inherent measurement noises, thereby causing human--robot conflicts in direction or parking location. According to the aforementioned scenario of the human--robot co-carrying task, we expect to design a continuous cooperation control framework, including reference generator and low-level controller, such that: (i) For the reference generator, when the predicted reference motion for the robot has errors, the predicted human motion can be corrected using admittance control based on the conflict measured by interaction forces. (ii) For the low-level controller of the robot, passivity is relaxed in a controlled manner, while simultaneously compensating energy, minimizing tracking errors, and regulating power flow to ensure both task performance and safe human--robot interaction.
\subsection{Preliminaries}
\begin{definition} 
\label{def1}
\cite{c38} A dynamic system is strictly passive with respect to the pair of input $\mathbf{u}$ and output $\mathbf{y}$ if there 
is a positive definite storage function $S\left( \mathbf{z} \right)$ and a positive definite dissipative function $D\left( \mathbf{z} \right)$ such that the following relationship holds for all $t\ge 0$: 
\begin{equation}
\label{eq1}
S\left( \mathbf{z}\left( t \right) \right)-S\left( \mathbf{z}\left( 0 \right) \right)= \int\limits_{0}^{t}{{{\mathbf{y}}^{\top}}\left( \tau  \right)\mathbf{u}\left( \tau  \right)d\tau -}\int\limits_{0}^{t}{D\left( \mathbf{z}\left( \tau  \right) \right)d\tau },
\end{equation}
in which $\mathbf{z}$ is the state vector.
\end{definition}
\begin{lemma}
\label{lm1}
\cite{c39} The following inequality holds for any real numbers $z_k$, $k=1,2,\ldots,n$, and for $0<r<1$:
\begin{equation}
\label{eq2}
\sum\limits_{k=1}^{n}{{{\left| {{z}_{k}} \right|}^{r}}}\ge {{\left( \sum\limits_{k=1}^{n}{\left| {{z}_{k}} \right|} \right)}^{r}}
\end{equation}
\end{lemma}

\section{Proposed Methodology}
\begin{figure*}[!t]
\centering
\includegraphics[scale=0.46]{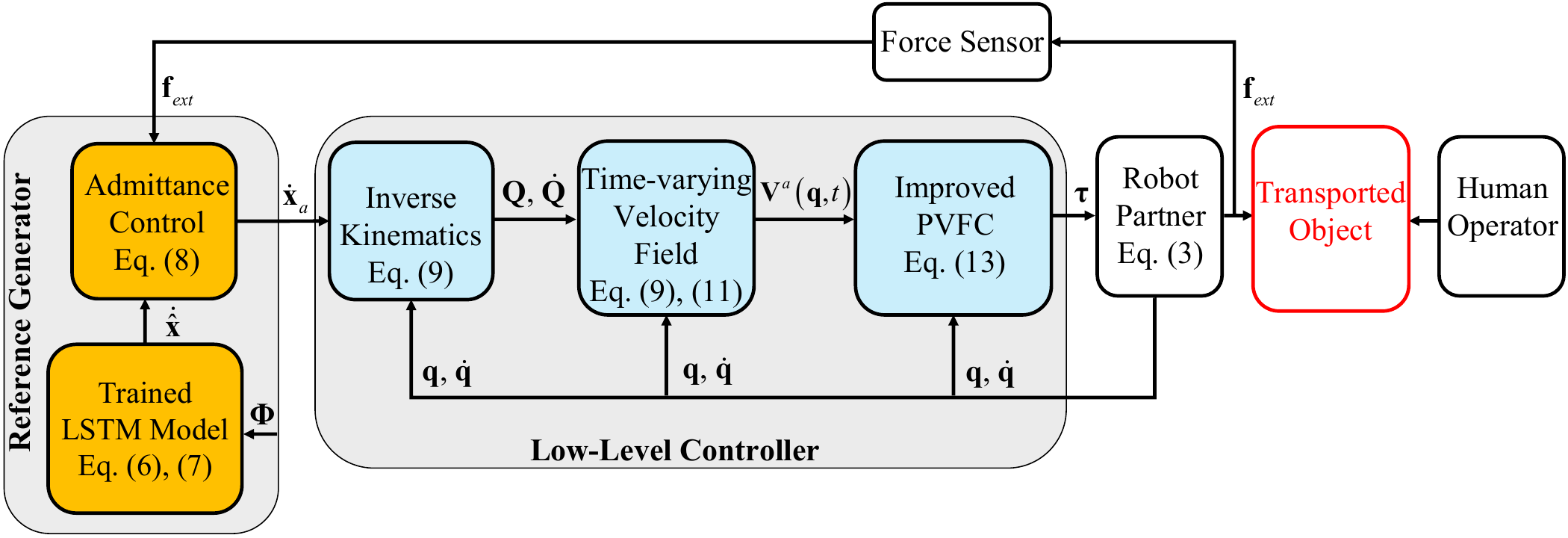}
\caption{Proposed cooperation control framework for human--robot co-carrying tasks, including reference generator and low-level controller. For the reference generator, a deep LSTM model is trained to predict human motion $\left( \mathbf{\hat{x}},\mathbf{\dot{\hat{x}}} \right)$ according to the previously measured motion data $\left( \mathbf{\Phi } \right)$. This predicted motion is corrected in the event of prediction errors thanks to admittance control 
with external forces $\left( {{\mathbf{f}}_{ext}}\right)$ which are measured by a force sensor. According to the output of the reference generator $\left( {{{\mathbf{\dot{x}}}}_{a}}\right)$ and robot's states $\left( \mathbf{q},\,\,\mathbf{\dot{q}} \right)$, the reference trajectory $\left( \mathbf{Q},\,\,\mathbf{\dot{Q}} \right)$ is encoded by a time-varying velocity field $\left( {{\mathbf{V}}^{a}}\left( {{\mathbf{q}}} \right),t \right)$ of the low-level controller. Thereupon, control torques for the robot's actuators $\left( \pmb{\tau } \right)$ are produced based on improved PVFC. }
\label{f2}
\end{figure*}
\subsection{System Description}
The dynamics of a fully actuated $n$-link robot manipulator is presented as,
\begin{equation}
\label{eq4}
\mathbf{M}\left( \mathbf{q} \right)\mathbf{\ddot{q}}+\mathbf{C}\left( \mathbf{q},\mathbf{\dot{q}} \right)\mathbf{\dot{q}}=\pmb{\tau }+{{\pmb{\tau }}_{ext}},
\end{equation}
in which $\mathbf{q}={{\left[ \begin{matrix}
   {{q}_{1}} & {{q}_{2}} & \ldots  & {{q}_{n}}  \\
\end{matrix} \right]}^{\top}}\in {{\mathbb{R}}^{{n}}}$ is the joint displacement vector, $\mathbf{M}\left( \mathbf{q} \right)$ and $\mathbf{C}\left( \mathbf{q},\mathbf{\dot{q}} \right)\in {{\mathbb{R}}^{{n\times n}}}$ denote the inertia, Coriolis and centrifugal matrices, respectively. $\pmb{\tau }$ and ${{\pmb{\tau }}_{ext}}\in {{\mathbb{R}}^{{n}}}$ are the control torque acting at the actuators and external torque from the interaction with the human operator, respectively. In the present study, we omit the gravitational vector in \eqref{eq4} to simplify the design and analysis of the cooperation control framework. By employing gravitational vector compensation methods, this term can be eliminated without compromising the generality of our study.

To guarantee the passivity of the system, the robot manipulator model in \eqref{eq4} is augmented with a fictitious flywheel state that serves as a fictitious energy storage component. The dynamics of the fictitious flywheel is defined as:
\begin{equation}
\label{eq5}
{{m}_{f}}{{\ddot{q}}_{f}}={{\tau }_{f}},
\end{equation}
where ${{m}_{f}}$ and ${{q}_{f}}$ represent the mass and angular position of the flywheel, ${{\tau }_{f}}$ is the control torque exerted on the flywheel. By combining the dynamic model of the robot manipulator in \eqref{eq4} and fictitious flywheel in \eqref{eq5}, an augmented system is given as,
\begin{equation}
\label{eq6}
{{\mathbf{M}}^{a}}\left( {{\mathbf{q}}^{a}} \right){{\mathbf{\ddot{q}}}^{a}}+{{\mathbf{C}}^{a}}\left( {{\mathbf{q}}^{a}},{{{\mathbf{\dot{q}}}}^{a}} \right){{\mathbf{\dot{q}}}^{a}}={{\pmb{\tau }}^{a}}+\pmb{\tau }_{ext}^{a},
\end{equation}
in which ${{\mathbf{q}}^{a}}={{\left[ \begin{matrix}
   {{q}_{1}} & {{q}_{2}} & \ldots  & \begin{matrix}
   {{q}_{n}} & {{q}_{f}}  \\
\end{matrix}  \\
\end{matrix} \right]}^{\top}}\in {{\mathbb{R}}^{n+1}}$ represents the joint angle vector, ${{\pmb{\tau }}^{a}}={{\left[ \begin{matrix}
   {{\pmb{\tau }}^{\top}} & {{\tau }_{f}}  \\
\end{matrix} \right]}^{\top}}$  and $\pmb{\tau }_{ext}^{a}={{\left[ \begin{matrix}
   \pmb{\tau }_{ext}^{\top} & 0  \\
\end{matrix} \right]}^{\top}}$ are the augmented control input and external force vector, respectively. ${{\mathbf{M}}^{a}}\left( {{\mathbf{q}}^{a}} \right)\,=\left[ \begin{matrix}
   \mathbf{M}\left( \mathbf{q} \right) & {{\mathbf{0}}_{n\times 1}}  \\
   {{\mathbf{0}}_{1\times n}}  & {{m}_{f}}  \\
\end{matrix} \right]$
and ${{\mathbf{C}}^{a}}\left( {{\mathbf{q}}^{a}},{{{\mathbf{\dot{q}}}}^{a}} \right)=\left[ \begin{matrix}
   \mathbf{C}\left( \mathbf{q},\mathbf{\dot{q}} \right) & {{\mathbf{0}}_{n\times 1}}  \\
   {{\mathbf{0}}_{1\times n}} & 0  \\
\end{matrix} \right]$  are the augmented inertia matrix and Coriolis matrix, respectively. ${{\mathbf{M}}^{a}}$ denotes a positive-definite symmetric matrix and $\left( {{{\mathbf{\dot{M}}}}^{a}}\left( {{\mathbf{q}}^{a}} \right)-2{{\mathbf{C}}^{a}}\left( {{\mathbf{q}}^{a}},{{{\mathbf{\dot{q}}}}^{a}} \right) \right)$ is a skew-symmetric matrix.
\subsection{Cooperation Control Framework Design}

In this section, a cooperation control framework with two parts is designed for collaboratively transporting a rigid object, as shown in Fig. \ref{f2}, including a reference generator and a low-level controller. The reference generator utilizes a deep long short-term memory (LSTM) network and admittance control to determine the robot's desired motion. Simultaneously, the low-level controller generates control torques to ensure synchronized human--robot motion while achieving safe interaction.
\subsubsection{Reference Generator}
\begin{figure}[!t]
\centering
\includegraphics[scale=0.25]{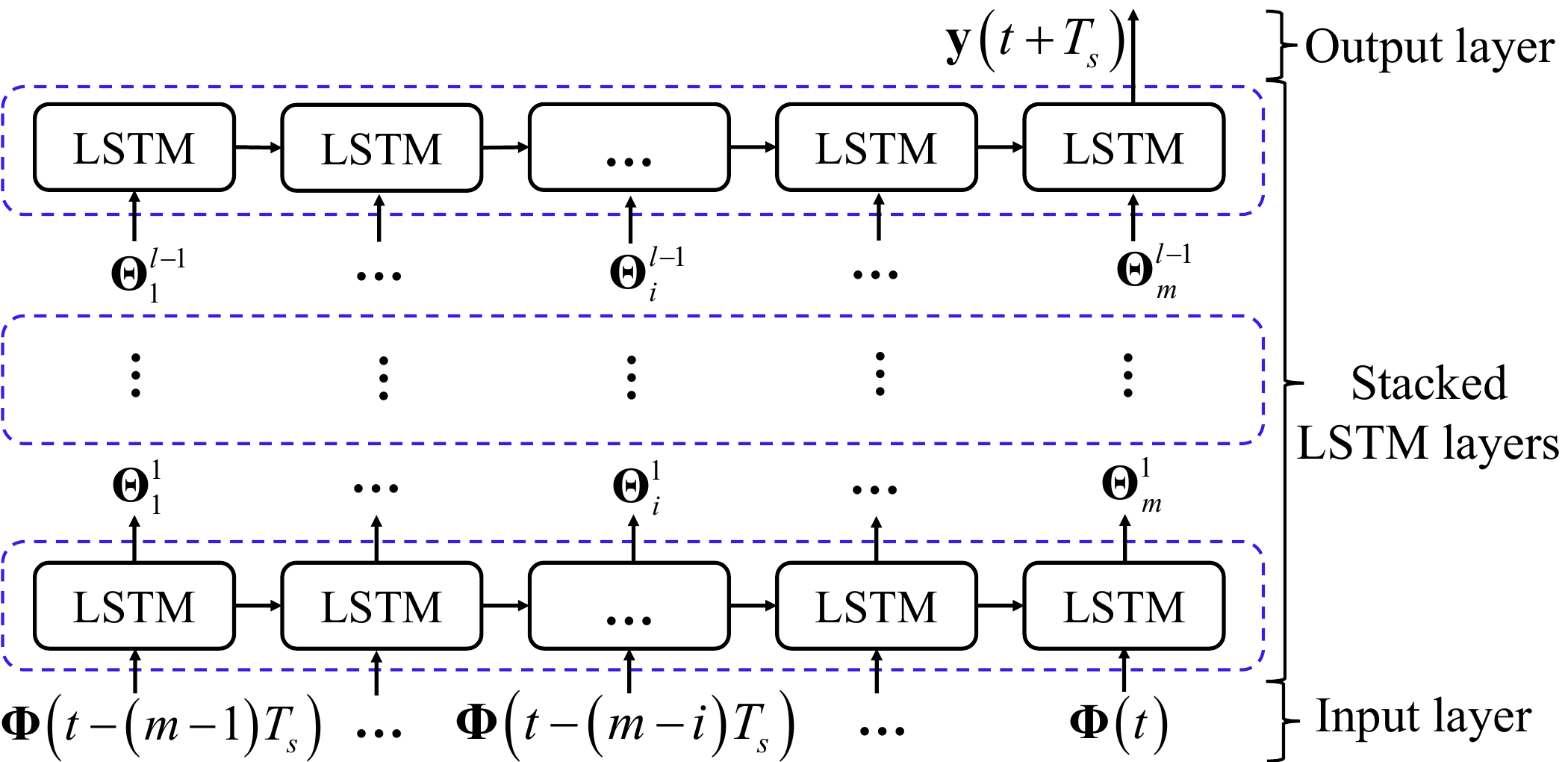}
\caption{Deep LSTM architecture for motion prediction. The input layer is established by the interaction motion of the collaborative co-carrying tasks, encompassing $m$ vectors $\mathbf{\Phi }\left( t-\left( m-i \right)T_s \right)$, $i=1,2,\ldots,m$. The stacked LSTM layers are designed by sequentially stacking multiple LSTM networks, where the output of each preceding network serves as the input for the subsequent network. The output layer $\left(\mathbf{y}\left( t+T_s \right)\right)$ can calculate the future reference position and velocity for the robot partner.}
\label{f3}
\end{figure}

Regarding the reference generator, an LSTM model-based data-driven approach is initially introduced to predict human motion. Leveraging recurrent connections with hidden states to retain memory of previous steps, LSTM networks are well-suited for handling sequential data of the co-carrying tasks. Furthermore, thanks to the cell state and four-gate structure, LSTM networks learn long-time dependencies and relieve gradient vanishing/ exploding issues in the traditional recurrent neural networks \cite{cRNN}. To enhance generalization ability and predictive performance, as pointed out in  \cite{cD}, a deep LSTM architecture is proposed in Fig. \ref{f3} with input, stacked LSTM, and output layers. 

The input layer of the deep LSTM Model in Fig. \ref{f3} is a sequence of $m$ vectors $\mathbf{\Phi }\left( t-\left( m-i \right)T_s \right)$, $i=1,2,\ldots,m$ at time $t$ with the sampling time $T_s$, in which each vector $\mathbf{\Phi }\left( t-\left( m-i \right)T_s \right)$ contains position and velocity on x and y-axes. In the present paper, we utilize coupled motion data, which is identified by the end-effector of the robot in collaboration, to reduce complexity for the human and dependence on visual sensors or electromyography. 

The stacked LSTM layer in Fig. 3 is constructed by stacking $l\ge2$ hidden LSTM layers. Specifically, the output of the former LSTM layer is the input of the latter LSTM layer.  Each LSTM layer is generated by $m$ recurrently connected LSTM cells corresponding to $m$ input vectors. According to \cite{cLSTM}, the $i$-th LSTM cell of the $j$-th layer with $i=1,2,\ldots,m$ and $j=1,2,\ldots,l$ is established by a input vector $\mathbf{\Theta }_{i}^{j}$ and a four-gate structure, including forget gate $\mathbf{f}_{i}^j$, input gate $\mathbf{i}_{i}^j$, cell gate $\mathbf{\tilde{c}}_{i}^j$, and output gate $\mathbf{o}_{i}^j$, as follows: 
\begin{equation}
\left\{ \begin{aligned}
  & {{\mathbf{f}}_{i}^j}=\sigma \left( {{\mathbf{W}}_\mathbf{f}^j}\left[ {{\mathbf{h}}_{i-1}^j},\mathbf{\Theta }_{i}^{j} \right]+{{\mathbf{b}}_\mathbf{f}^j} \right) \\ 
 & {{\mathbf{i}}_{i}^j}=\sigma \left( {{\mathbf{W}}_\mathbf{i}^j}\left[ {{\mathbf{h}}_{i-1}^j},\mathbf{\Theta }_{i}^{j} \right]+{{\mathbf{b}}_\mathbf{i}^j} \right) \\ 
 & {{{\mathbf{\tilde{c}}}}_{i}^j}=\tanh \left( {{\mathbf{W}}_\mathbf{c}^j}\left[ {{\mathbf{h}}_{i-1}^j},\mathbf{\Theta }_{i}^{j} \right]+{{\mathbf{b}}_\mathbf{c}^j} \right) \\ 
 & {{\mathbf{o}}_{i}^j}=\sigma \left( {{\mathbf{W}}_\mathbf{o}^j}\left[ {{\mathbf{h}}_{i-1}^j},\mathbf{\Theta }_{i}^{j} \right]+{{\mathbf{b}}_\mathbf{o}^j} \right) \\ 
 & {{\mathbf{c}}_{i}^j}={{\mathbf{f}}_{i}^j}{{\mathbf{c}}_{i-1}^j}+{{\mathbf{i}}_{i}^j}{{{\mathbf{\tilde{c}}}}_{i}^j} \\ 
 & {{\mathbf{h}}_{i}^j}={{\mathbf{o}}_{i}^j}\tanh \left( {{\mathbf{c}}_{i}^j} \right) \\ 
\end{aligned} \right.
\label{eq6.1}
\end{equation}
in which $\mathbf{W}_\mathbf{f}^j, \mathbf{W}_\mathbf{i}^j, \mathbf{W}_\mathbf{c}^j, \mathbf{W}_\mathbf{o}^j$ and $\mathbf{b}_\mathbf{f}^j, \mathbf{b}_\mathbf{i}^j, \mathbf{b}_\mathbf{c}^j, \mathbf{b}_\mathbf{o}^j$ are weight matrices and bias vectors of the forget gate, input gate, cell gate, and output gate, respectively. $\mathbf{c}_{i-1}^j$, $\mathbf{h}_{i-1}^j$ and  $\mathbf{c}_{i}^j$, $\mathbf{h}_{i}^j$ are the cell state vector and hidden state vector of the previous network and the current network. Note that $\mathbf{\Theta }_{i}^{0}=\mathbf{\Phi }\left( t-\left( m-i \right)T_s \right)$ are utilized for the first LSTM layer. $\sigma \left( \right)$ and $\tanh \left( \right)$ represent the sigmoid function and hyperbolic tangent function.

For the output layer, the hidden vector of the $m$-th cell of the last LSTM layer, denoted by $\left( {{\mathbf{h}}_{m}^l} \right)$, is utilized to generate the following motion prediction.
\begin{equation}
\mathbf{y}\left( t+T_s \right)={{\mathbf{W}}_{y}}{{\mathbf{h}}_{m}^l}+{{\mathbf{b}}_{y}},
\label{eq6.2}
\end{equation}
where $\mathbf{y}\left( t+T_s \right)={{\left[ \begin{matrix}
   {{{\mathbf{\hat{x}}}}^{\top}} & {{{\mathbf{\dot{\hat{x}}}}}^{\top}}  \\
\end{matrix} \right]}^{\top}}$. ${{\mathbf{W}}_{y}}$ and ${{\mathbf{b}}_{y}}$ are weight matrix and bias vector of output layer. 

However, the predicted human motion can be inaccurate because of the diverse human intentions in adapting to unstructured workspaces and inherent measurement noises. Therefore, a fixed admittance control with mass--damper (MD) features integrating predicted motion $\left({{{\mathbf{\hat{x}}}}}\right)$ is presented to re-plan the desired co-carrying trajectory for the robot partner according to external forces exerted by the human $\left({{\mathbf{f}}_{ext}}\right)$ in the presence of prediction errors.
\begin{equation}
{{\mathbf{M}}_{d}}\left( {{{\mathbf{\ddot{x}}}}_{a}}-{{{\mathbf{\ddot{\hat{x}}}}} }\right)+{{\mathbf{D}}_{d}}\left( {{{\mathbf{\dot{x}}}}_{a}}-{{{\mathbf{\dot{\hat{x}}}}}} \right)={{\mathbf{f}}_{ext}},
\label{eq7}
\end{equation}
where ${{\mathbf{M}}_{d}}$ and ${{\mathbf{D}}_{d}}$ are diagonal matrices of inertia and damping terms, respectively. ${{\mathbf{x}}_{a}}$ represents the desired trajectory obtained from \eqref{eq7} and ${{{\mathbf{\hat{x}}}}}$ is the reference motion identified by the deep LSTM structure. ${{\mathbf{f}}_{ext}}$ denotes the external force that is applied by the human. Based on \eqref{eq7}, the characteristics of the admittance control are stated as follows. When the deep LSTM predicts the reference motion toward human intention, the human--robot interaction force $\left({{\mathbf{f}}_{ext}}\right)$ is very small or ideally zero. Specifically, by considering diagonal positive-definite matrices $\left( {{\mathbf{M}}_{d}},{{\mathbf{D}}_{d}} \right)$ in \eqref{eq7}, we can see that ${{\mathbf{\dot{x}}}_{a}}=\mathbf{\dot{\hat{x}}}+\left( {{{\mathbf{\dot{x}}}}_{a}}\left( 0 \right)-\mathbf{\dot{\hat{x}}}\left( 0 \right) \right){{e}^{-\mathbf{M}_{d}^{-1}{{\mathbf{D}}_{d}}t}}$ when ${{\mathbf{f}}_{ext}}=\mathbf{0}$. As a result, the desired motion ${{{\mathbf{\dot{x}}}}_{a}}$ converges to the reference motion ${{{\mathbf{\dot{\hat{x}}}}}}$ as well as human motion intention. On the contrary, the desired co-carrying motion for the robot ${{{\mathbf{\dot{x}}}}_{a}}$ is re-generated according to the interaction force $\left({{\mathbf{f}}_{ext}}\right)$ and the reference motion ${{{\mathbf{\dot{\hat{x}}}}}}$. In this case, the human tends to exert forces $\left( {{\mathbf{f}}_{ext}}\ne \mathbf{0} \right)$ to accelerate or decelerate the robot motion, thereby enabling the desired motion ${{\mathbf{x}}_{a}}$ to more closely align with human motion intention.

\subsubsection{Low-Level Controller}For the low-level controller, the output of the reference generator $\left( {{{\mathbf{\dot{x}}}}_{a}} \right)$ is encoded by the time-varying velocity field of the augmented system \eqref{eq6} $\left( {{\mathbf{V}}^{a}}\left( {{\mathbf{q}}},t \right)={{\left[ \begin{matrix}
   {{\mathbf{V}}^{\top}}\left( \mathbf{q},t \right) & {{V}_{f}}\left( \mathbf{q},t \right)  \\
\end{matrix} \right]}^{\top}} \right)$ to aim to design control torques for the robot. Firstly, the time-varying velocity field $\left( \mathbf{V}\left( \mathbf{q},t \right) \right)$ of the robot manipulator in \eqref{eq4} is created as 
\begin{equation}
\left\{ \begin{aligned}
  & \mathbf{\dot{Q}}\left( t \right)={{\mathbf{J}}^{+}}\left( \mathbf{q} \right){{{\mathbf{\dot{x}}}}_{a}}\left( t \right) \\ 
 & \mathbf{V}\left( \mathbf{q},t \right)=\mathbf{\dot{Q}}\left( t \right)-{{\mathbf{K}}_{1}}\left( \mathbf{q}\left( t \right)-\mathbf{Q}\left( t \right) \right), \\ 
\end{aligned} \right.
\label{eq8}
\end{equation}
in which ${{\mathbf{J}}^{+}}\left( \mathbf{q} \right)={{\left( {{\mathbf{J}}^{\top}}\left( \mathbf{q} \right)\mathbf{J}\left( \mathbf{q} \right) \right)}^{-1}}{{\mathbf{J}}^{\top}}\left( \mathbf{q} \right)$ with the Jacobian matrix $\mathbf{J}\left( \mathbf{q} \right)$ and $\mathbf{Q}\left( t \right)$ is the reference angular position. ${{\mathbf{K}}_{1}}$ denotes the positive-definite control parameter matrix. According to \cite{c18}, the desired time-varying velocity field 
$\left( {{\mathbf{V}}^{a}}\left( {{\mathbf{q}}},t \right)={{\left[ \begin{matrix}
   {{\mathbf{V}}^{\top}}\left( \mathbf{q},t \right) & {{V}_{f}}\left( \mathbf{q},t \right)  \\
\end{matrix} \right]}^{\top}} \right)$ satisfies consistency and the following conservation of kinetic energy principle.
\begin{equation}
\begin{aligned}
  & {{k}^{a}}\left( {{\mathbf{q}}^{a}},\,\,{{\mathbf{V}}^{a}}\left( \mathbf{q},t \right) \right)=\frac{1}{2}{{\mathbf{V}}^{a}}^\top\left( \mathbf{q},t \right){{\mathbf{M}}^{a}}\left( {{\mathbf{q}}^{a}} \right){{\mathbf{V}}^{a}}\left( \mathbf{q},t \right) \\ 
 & \,\,\,\,\,\,\,\,\,\,\,\,\,\,\,\,=\frac{1}{2}{{\mathbf{V}}^{\top}}\left( \mathbf{q},t \right)\mathbf{M}\left( \mathbf{q} \right)\mathbf{V}\left( \mathbf{q},t \right)+\frac{1}{2}{{m}_{f}}V_{f}^{2}\left( \mathbf{q},t \right) \\ 
 & \,\,\,\,\,\,\,\,\,\,\,\,\,\,\,\,={{E}^{a}}>0,
\end{aligned}
\label{eq10}
\end{equation}
in which ${{k}^{a}}\left( {{\mathbf{q}}^{a}},\,\,{{\mathbf{V}}^{a}}\left( \mathbf{q},t \right) \right)$ is the kinetic energy at the desired velocity field. From \eqref{eq10}, the desired velocity field ${{V}_{f}}\left( \mathbf{q},t \right)$  of the fictitious flywheel is calculated as:
\begin{equation}
{{V}_{f}}\left( \mathbf{q},t \right)=\sqrt{\frac{2}{{{m}_{f}}}\left( {{E}^{a}}-\frac{1}{2}{{\mathbf{V}}^{\top}}\left( \mathbf{q},t \right)\mathbf{M}\left( \mathbf{q} \right)\mathbf{V}\left( \mathbf{q},t \right) \right)}
\label{eq9}
\end{equation}
Note that ${{E}^{a}}$ denotes a sufficiently large positive constant, which ensures the term under the square root is positive. 
\begin{remark}
\label{r1}
According to the paper \cite{c18}, the velocity field focuses on minimizing the error between actual position and contour rather than attempting to leave the contour to synchronize with the desired position as the conventional timed trajectory method \cite{c18.4}. Utilizing the velocity field concept in the human--robot co-carrying tasks enables the robot partner to adhere to the co-carrying contour instead of abruptly advancing toward the target position. Specifically, the gradient vector field of the distance between $\mathbf{q}\left( t \right)$ and $\mathbf{Q}\left( t \right)$ in \eqref{eq8} adjusts the velocity vector based on the robot's current state. Furthermore, the expression \eqref{eq8} also encodes the time-varying trajectories, facilitating the robot to accommodate varying human intentions in adapting to unstructured environments.
\end{remark}
 In addition, the actual kinetic energy $\left({{k}^{a}}\left( {{\mathbf{q}}^{a}},\,\,{{{\mathbf{\dot{q}}}}^{a}} \right)\right)$ of the augmented system is computed as,
\begin{equation}
\begin{aligned}
  & {{k}^{a}}\left( {{\mathbf{q}}^{a}},\,\,{{\mathbf{\dot{q}}}^{a}} \right)=\frac{1}{2}{{\mathbf{\dot{q}}}^{a\top}}{{\mathbf{M}}^{a}}\left( {{\mathbf{q}}^{a}} \right){{{\mathbf{\dot{q}}}}^{a}} \\ 
 & \,\,\,\,\,\,\,\,\,\,\,\,\,\,\,\,\,\,\,\,\,\,\,\,\,\,\,\,\,\,=\frac{1}{2}{{{\mathbf{\dot{q}}}}^{\top}}\mathbf{M}\left( \mathbf{q} \right)\mathbf{\dot{q}}+\frac{1}{2}{{m}_{f}}\dot{q}_{f}^{2}. \\ 
\end{aligned}
\label{eq11}
\end{equation}

An improved time-varying PVFC is designed to guarantee the flexible passivity of the closed-loop system. The approach in \cite{c20} is inherited in order to control the power flow, such that it enhances the safety in human--robot collaborative tasks. Furthermore, we newly propose a fractional exponent control term to compensate for the kinetic energy of the augmented system in a finite time interval. Based on \eqref{eq7}--\eqref{eq11}, the improved PVFC with a fractional exponent energy compensation control term  is developed for the ${{\pmb{\tau }}^{a}}\left( {{\mathbf{q}}^{a}},{{{\mathbf{\dot{q}}}}^{a}},t \right)$ of the augmented system in \eqref{eq6} as, 
\begin{equation}
\begin{aligned}
  & {{\pmb{\tau }}^{a}}\left( {{\mathbf{q}}^{a}},{{{\mathbf{\dot{q}}}}^{a}},t \right)=\left[ \mathbf{G}\left( {{\mathbf{q}}^{a}},{{{\mathbf{\dot{q}}}}^{a}},t \right)+\mathbf{R}\left( {{\mathbf{q}}^{a}},{{{\mathbf{\dot{q}}}}^{a}},t \right) \right]{{{\mathbf{\dot{q}}}}^{a}} \\ 
 & \,\,\,\,\,\,\,\,\,\,\,\,\,\,\,\,\,\,\,\,\,\,\,\,\,\,\,\,\,\,\,\,\,\,\,\,\,\,\,\,\,\,\,\,\,\,\,\,\,\,\,\,\,\,\,\,\,\,\,\,\,\,\,\,\,\,\,\,\,\,\,\,-\mathbf{S}\left( {{\mathbf{q}}^{a}},{{{\mathbf{\dot{q}}}}^{a}} \right){{\left[ {{{\mathbf{\dot{q}}}}^{a}} \right]}^{\frac{{{r}_{1}}}{{{r}_{2}}}}}, \\ 
\end{aligned}
\label{eq12}
\end{equation}
in which $\mathbf{S}\left( {{\mathbf{q}}^{a}},{{{\mathbf{\dot{q}}}}^{a}} \right){{\left[ {{{\mathbf{\dot{q}}}}^{a}} \right]}^{\frac{{{r}_{1}}}{{{r}_{2}}}}}$  is utilized to compensate for the kinetic energy of the augmented system within a finite time interval. Two skew-symmetric matrices $\mathbf{G}\left( {{\mathbf{q}}^{a}},{{{\mathbf{\dot{q}}}}^{a}},t \right)$, $\mathbf{R}\left( {{\mathbf{q}}^{a}},{{{\mathbf{\dot{q}}}}^{a}},t \right)$, and a diagonal matrix $\mathbf{S}\left( {{\mathbf{q}}^{a}},{{{\mathbf{\dot{q}}}}^{a}} \right)$ are designed in the following formulation.
\begin{equation}
\left\{ \begin{aligned}
  & \mathbf{G}\left( {{\mathbf{q}}^{a}},{{{\mathbf{\dot{q}}}}^{a}},t \right):=\frac{1}{2{{E}^{a}}}\left( \mathbf{w}{{\mathbf{P}}^{\top}}-\mathbf{P}{{\mathbf{w}}^{\top}} \right) \\ 
 & \mathbf{R}\left( {{\mathbf{q}}^{a}},{{{\mathbf{\dot{q}}}}^{a}},t \right):=\kappa \left( \mathbf{P}{{\mathbf{p}}^{\top}}-\mathbf{p}{{\mathbf{P}}^{\top}} \right) \\ 
 & \mathbf{S}\left( {{\mathbf{q}}^{a}},{{{\mathbf{\dot{q}}}}^{a}} \right):=s\left( {{k}^{a}}\left( {{\mathbf{q}}^{a}},{{{\mathbf{\dot{q}}}}^{a}} \right)-k_{d}^{a} \right){{\mathbf{K}}_{2}}, \\ 
\end{aligned} \right.
\label{eq13}
\end{equation}
where $\kappa$ is the control parameter and $k_{d}^{a}$ is the desired kinetic energy level of the augmented system. Furthermore, to avoid clutter, $\mathbf{w}=\mathbf{w}\left( {{\mathbf{q}}^{a}},{{{\mathbf{\dot{q}}}}^{a}},t \right)$, $\mathbf{P}=\mathbf{P}\left( {{\mathbf{q}}^{a}},t \right)$, and  $\mathbf{p}=\mathbf{p}\left( {{\mathbf{q}}^{a}},{{{\mathbf{\dot{q}}}}^{a}} \right)$ are utilized in the paper. The inverse dynamics with the desired time-varying velocity field $\left(\mathbf{w}\right)$, desired momentum $\left(\mathbf{P}\right)$, and actual momentum $\left(\mathbf{p}\right)$ of the augmented system in \eqref{eq6}, as well as the diagonal positive-definite matrix $\left({{\mathbf{K}}_{2}}\right)$ are defined as:
\begin{equation}
\left\{ \begin{aligned}
  & \mathbf{w}={{\mathbf{M}}^{a}}\left( {{\mathbf{q}}^{a}} \right){{{\mathbf{\dot{V}}}}^{a}}\left( \mathbf{q},t \right)+{{\mathbf{C}}^{a}}\left( {{\mathbf{q}}^{a}},{{{\mathbf{\dot{q}}}}^{a}} \right){{\mathbf{V}}^{a}}\left( \mathbf{q},t \right) \\ 
 & \mathbf{P}={{\mathbf{M}}^{a}}\left( {{\mathbf{q}}^{a}} \right){{\mathbf{V}}^{a}}\left( \mathbf{q},t \right) \\ 
 & \mathbf{p}={{\mathbf{M}}^{a}}\left( {{\mathbf{q}}^{a}} \right){{{\mathbf{\dot{q}}}}^{a}} \\ 
 & {{\mathbf{K}}_{2}}=\text{diag}\left( {{k}_{21}},{{k}_{22}},\ldots ,{{k}_{2n}},{{k}_{2f}} \right), \\ 
\end{aligned} \right.
\label{eq14}
\end{equation}
where the $i$-th term of the differential of the time-varying velocity field is computed as follows,
\begin{equation}
\dot{V}_{i}^{a}\left( \mathbf{q},t \right)=\sum\limits_{j=1}^{n+1}{\frac{\partial V_{i}^{a}\left( \mathbf{q},t \right)}{\partial q_{j}^{a}}\dot{q}_{j}^{a}}+\frac{\partial V_{i}^{a}\left( \mathbf{q},t \right)}{\partial t}. 
\label{eq15}
\end{equation}
Additionally, the fractional exponent of the augmented velocity in \eqref{eq12} is defined by:
\begin{equation}
\begin{aligned}
  & {{\left[ {{{\mathbf{\dot{q}}}}^{a}} \right]}^{\frac{{{r}_{1}}}{{{r}_{2}}}}}=\left[ \begin{matrix}
   {\mathop{\rm sgn}} \left( {{{\dot{q}}}_{1}} \right){{\left| {{{\dot{q}}}_{1}} \right|}^{\frac{{{r}_{1}}}{{{r}_{2}}}}} & {\mathop{\rm sgn}} \left( {{{\dot{q}}}_{2}} \right){{\left| {{{\dot{q}}}_{2}} \right|}^{\frac{{{r}_{1}}}{{{r}_{2}}}}} & \ldots   \\
\end{matrix} \right. \\ 
 & \,\,\,\,\,\,\,\,\,\,\,\,\,\,\,\,\,\,\,\,\,\,\,\,\,\,\,\,\,\,\,\,\,\,\,\,\,\,\,\,\,{{\left. \begin{matrix}
   {\mathop{\rm sgn}} \left( {{{\dot{q}}}_{n}} \right){{\left| {{{\dot{q}}}_{n}} \right|}^{\frac{{{r}_{1}}}{{{r}_{2}}}}} & {\mathop{\rm sgn}} \left( {{{\dot{q}}}_{f}} \right){{\left| {{{\dot{q}}}_{f}} \right|}^{\frac{{{r}_{1}}}{{{r}_{2}}}}}  \\
\end{matrix} \right]}^{T}}, \\ 
\end{aligned}
\label{eq16}
\end{equation}
with ${{r}_{1}},\,{{r}_{2}}$ $\left( {{r}_{1}}<{{r}_{2}} \right)$ are  positive integers. The smooth saturation function in \eqref{eq13} is designed to compensate for the kinetic energy of the augmented system.
\begin{equation}
s\left( e \right)=\left\{ \begin{aligned}
  & {{\eta }_{\min }}\,\,\,\,\,\,\,\,\,\,\,\,\,\,\,\,\,\,\,\,\,\,\,\,\,\,\,\,\,\,\,\,\,\,\,\,\,\,\,\,\,\,\,\,\,\,\,\,\,\,\,\,\,\,\,\,\,\,\,\,\,\,\,\,\,e<{{\delta }_{1}} \\ 
 & \frac{{{\eta }_{\min }}}{2}\left( 1-\cos \left( \frac{\pi }{{{\delta }_{1}}}e \right) \right)\,\,\,\,\,\,\,\,\,\,{{\delta }_{1}}\le e<0\, \\ 
 & 0\,\,\,\,\,\,\,\,\,\,\,\,\,\,\,\,\,\,\,\,\,\,\,\,\,\,\,\,\,\,\,\,\,\,\,\,\,\,\,\,\,\,\,\,\,\,\,\,\,\,\,\,\,\,\,\,\,\,\,\,\,\,\,\,\,\,\,\,\,\,\,\,\,\,\,\,e=0\\ 
 & \frac{{{\eta }_{\max }}}{2}\left( 1-\cos \left( \frac{\pi }{{{\delta }_{2}}}e \right) \right)\,\,\,\,\,\,\,\,\,0<e\le {{\delta }_{2}}\, \\ 
 & {{\eta }_{\max }}\,\,\,\,\,\,\,\,\,\,\,\,\,\,\,\,\,\,\,\,\,\,\,\,\,\,\,\,\,\,\,\,\,\,\,\,\,\,\,\,\,\,\,\,\,\,\,\,\,\,\,\,\,\,\,\,\,\,\,\,\,\,\,\,\,\,e>{{\delta }_{2}} \\ 
\end{aligned} \right.
\label{eq17}
\end{equation}
where $e={{k}^{a}}\left( {{\mathbf{q}}^{a}},{{{\mathbf{\dot{q}}}}^{a}} \right)-k_{d}^{a}$ and ${{\eta }_{\min }}<0$, ${{\eta }_{\max }}>0$, ${{\delta }_{2}}>0$, ${{\delta }_{1}}<0$ are constants.
\subsection{Main Results}
\begin{theorem}
\label{thm1} {\emph{(Passivity)}}
Leveraging the action of control commands in \eqref{eq12} for the augmented system in \eqref{eq6}, the kinetic energy of the system in \eqref{eq11} satisfies:
\begin{equation}
\begin{aligned}
  & \frac{d}{dt}{{k}^{a}}\left( {{\mathbf{q}}^{a}},{{{\mathbf{\dot{q}}}}^{a}} \right)=\pmb{\tau }_{ext}^{a\top}{{{\mathbf{\dot{q}}}}^{a}} \\ 
 & \,\,\,\,\,\,\,\,\,\,\,\,\,\,-s\left( {{k}^{a}}\left( {{\mathbf{q}}^{a}},{{{\mathbf{\dot{q}}}}^{a}} \right)-k_{d}^{a} \right){{{\mathbf{\dot{q}}}}^{a\top}}{{\mathbf{K}}_{2}}{{\left[ {{{\mathbf{\dot{q}}}}^{a}} \right]}^{\frac{{{r}_{1}}}{{{r}_{2}}}}}, \\ 
\end{aligned}
\label{eq18}
\end{equation}
in which ${{r}_{1}},\,{{r}_{2}}$ $\left( {{r}_{1}}<{{r}_{2}} \right)$ are positive integers and ${{\mathbf{K}}_{2}}$ is a positive-definite matrix. According to Definition \ref{def1}, the closed-loop system is strictly passive with respect to the pair of the velocity output $\left( {{{\mathbf{\dot{q}}}}^{a}} \right)$ and external force input $\left( \pmb{\tau }_{ext}^{a} \right)$ under the condition ${{k}^{a}}\left( {{\mathbf{q}}^{a}},{{{\mathbf{\dot{q}}}}^{a}} \right)-k_{d}^{a}\ge 0$.
\end{theorem}

\begin{theorem}
\label{thm2} {\emph{(Convergence time of kinetic energy)}}
Under the improved PVFC in \eqref{eq12} with positive integers $\left( {{r}_{1}},\,{{r}_{2}}\,\,\left( {{r}_{1}}<{{r}_{2}} \right) \right)$ and positive-definite matrix $\left( {{\mathbf{K}}_{2}} \right)$, the kinetic energy of the augmented system ${{k}^{a}}\left( {{\mathbf{q}}^{a}},{{{\mathbf{\dot{q}}}}^{a}} \right)$ converges to the desired one $k_{d}^{a}$ within a finite time interval in the absence of external disturbance $\left( \pmb{\tau }_{ext}^{a}=0 \right)$. The settling time of the system $\left( T \right)$ is bounded by,
\begin{equation}
T\le \max \left\{ {{T}_{1}},{{T}_{2}} \right\},
\label{eq19}
\end{equation}
where $T_1$ and $T_2$ are respectively the convergence times with ${{k}^{a}}\left( {{\mathbf{q}}^{a}}, {{{\mathbf{\dot{q}}}}^{a}} \right)\ge k_{d}^{a}$ and ${{k}^{a}}\left( {{\mathbf{q}}^{a}},{{{\mathbf{\dot{q}}}}^{a}} \right)<k_{d}^{a}$, which are specifically calculated in the next subsection.  
\end{theorem}

\begin{theorem}
\label{thm3} {\emph{(Stability)}}
By applying the control inputs of the improved time-varying PVFC in \eqref{eq8} and \eqref{eq12} and designing the positive-definite control parameters $\left( {{\mathbf{K}}_{1}},\,\,\kappa  \right)$,  the solutions
\begin{equation}
\left\{ \begin{aligned}
  & {{\mathbf{e}}_{\mathbf{q}}}=\mathbf{q}\left( t \right)-\mathbf{Q}\left( t \right)=\mathbf{0} \\ 
 & {{\mathbf{e}}_{\mathbf{v}}}={{{\mathbf{\dot{q}}}}^{a}}\left( t \right)-\alpha \left( t \right){{\mathbf{V}}^{a}}\left({{\mathbf{q}}^{a}}\left( t \right), t \right)=\mathbf{0} \\ 
\end{aligned} \right.
\label{eq20}
\end{equation}
to the augmented system in \eqref{eq6} is Lyapunov stable in the absence of external disturbance $\left( \pmb{\tau }_{ext}^{a}=0 \right)$. The $\alpha \left( t \right)$ term is defined such that:
\begin{equation}
\alpha \left( t \right)=\sqrt{\frac{{{k}^{a}}\left( {{\mathbf{q}}^{a}}\left( t \right),{{{\mathbf{\dot{q}}}}^{a}}\left( t \right) \right)}{{{E}^{a}}}}.
\label{eq21}
\end{equation}
\end{theorem}

\begin{corollary}
\label{lm3} {\emph{(Power flow)}}
When the external force $\left( \pmb{\tau }_{ext}^{a} \right)$ is applied by the human operator, the power flow into the human and robot is calculated by the augmented system in \eqref{eq6} and the control command in \eqref{eq12} as follows:  
\begin{equation}
\label{eq22}
{{P}_{r2h}}\,\,=-s\left( {{k}^{a}}\left( \mathbf{q},{{\mathbf{q}}^{a}} \right)-k_{d}^{a} \right){{\mathbf{\dot{q}}}^{aT}}{{\mathbf{K}}_{2}}{{\left[ {{{\mathbf{\dot{q}}}}^{a}} \right]}^{\frac{{{r}_{1}}}{{{r}_{2}}}}}-{{\dot{k}}^{a}},
\end{equation}
in which ${{P}_{r2h}}\,>0$  is the power flow from the robot to the human; meanwhile, ${{P}_{r2h}}\,<0$ denotes the power flowing from the human to the robot. By adjusting the control parameters $\left( {{\mathbf{K}}_{2}},{{r}_{1}},r_{2}^{{}}, {{\eta }_{\max }},{{\eta }_{\min }},{{\delta }_{1}},{{\delta }_{2}}\right)$ in \eqref{eq13} and \eqref{eq17}, we can regulate the power flow in the collaboration systems.

\end{corollary}
\subsection{Proof}
\begin{proof1}
\label{pt1}
Differentiating the kinetic energy in \eqref{eq11} and utilizing the augmented system in \eqref{eq6}, yields: 
\begin{equation}
\label{eq24}
\begin{aligned}
  & \frac{d}{dt}{{k}^{a}}\left( {{\mathbf{q}}^{a}},{{{\mathbf{\dot{q}}}}^{a}} \right) ={{{\mathbf{\dot{q}}}}^{a\top}}{{\mathbf{M}}^{a}}{{{\mathbf{\ddot{q}}}}^{a}}+\frac{1}{2}{{{\mathbf{\dot{q}}}}^{a\top}}{{{\mathbf{\dot{M}}}}^{a}}{{{\mathbf{\dot{q}}}}^{a}} \\ 
 & =\frac{1}{2}{{{\mathbf{\dot{q}}}}^{a\top}}{{{\mathbf{\dot{M}}}}^{a}}{{{\mathbf{\dot{q}}}}^{a}}+{{{\mathbf{\dot{q}}}}^{a\top}}\left( {{\pmb{\tau }}^{a}}+\pmb{\tau }_{ext}^{a}-{{\mathbf{C}}^{a}}\left( {{\mathbf{q}}^{a}},{{{\mathbf{\dot{q}}}}^{a}} \right){{{\mathbf{\dot{q}}}}^{a}} \right). \\ 
\end{aligned}
\end{equation}
Substituting the improved time-varying PVFC in \eqref{eq12} into \eqref{eq24}, we obtain,
\begin{equation}
\label{eq25}
\begin{aligned}
  & \frac{d}{dt}{{k}^{a}}\left( {{\mathbf{q}}^{a}},{{{\mathbf{\dot{q}}}}^{a}} \right)=\frac{1}{2}{{{\mathbf{\dot{q}}}}^{a\top}}\left( {{{\mathbf{\dot{M}}}}^{a}}-2{{\mathbf{C}}^{a}}\left( {{\mathbf{q}}^{a}},{{{\mathbf{\dot{q}}}}^{a}} \right) \right){{{\mathbf{\dot{q}}}}^{a}} \\ 
 & \,\,\,\,\,\,\,\,\,\,\,\,\,\,\,\,\,\,+{{{\mathbf{\dot{q}}}}^{a\top}}\mathbf{R}\left( {{\mathbf{q}}^{a}},{{{\mathbf{\dot{q}}}}^{a}},t \right){{{\mathbf{\dot{q}}}}^{a}}+{{{\mathbf{\dot{q}}}}^{a\top}}\mathbf{G}\left( {{\mathbf{q}}^{a}},{{{\mathbf{\dot{q}}}}^{a}},t \right){{{\mathbf{\dot{q}}}}^{a}} \\ 
 & \,\,\,\,\,\,\,\,\,\,\,\,\,\,\,\,\,\,+{{{\mathbf{\dot{q}}}}^{a\top}}\pmb{\tau }_{ext}^{a}-{{{\mathbf{\dot{q}}}}^{a\top}}\mathbf{S}\left( {{\mathbf{q}}^{a}},{{{\mathbf{\dot{q}}}}^{a}} \right){{\left[ {{{\mathbf{\dot{q}}}}^{a}} \right]}^{\frac{{{r}_{1}}}{{{r}_{2}}}}}, \\ 
\end{aligned}
\end{equation}
where ${{\mathbf{\dot{M}}}^{a}}-2{{\mathbf{C}}^{a}}\left( {{\mathbf{q}}^{a}},{{{\mathbf{\dot{q}}}}^{a}} \right),\,\,\mathbf{G}\left( {{\mathbf{q}}^{a}},{{{\mathbf{\dot{q}}}}^{a}},t \right),\,\,\mathbf{R}\left( {{\mathbf{q}}^{a}},{{{\mathbf{\dot{q}}}}^{a}},t \right)$ are skew-symmetric matrices; therefore, \eqref{eq25} becomes:
\begin{equation}
\label{eq26}
\begin{aligned}
  & \frac{d}{dt}{{k}^{a}}\left( {{\mathbf{q}}^{a}},{{{\mathbf{\dot{q}}}}^{a}} \right) ={{{\mathbf{\dot{q}}}}^{a\top}}\pmb{\tau }_{ext}^{a}-{{{\mathbf{\dot{q}}}}^{a\top}}\mathbf{S}\left( {{\mathbf{q}}^{a}},{{{\mathbf{\dot{q}}}}^{a}} \right){{\left[ {{{\mathbf{\dot{q}}}}^{a}} \right]}^{\frac{{{r}_{1}}}{{{r}_{2}}}}} \\ 
 & =\pmb{\tau }_{ext}^{a\top}{{{\mathbf{\dot{q}}}}^{a}}-s\left( {{k}^{a}}\left( {{\mathbf{q}}^{a}},{{{\mathbf{\dot{q}}}}^{a}} \right)-k_{d}^{a} \right){{{\mathbf{\dot{q}}}}^{a\top}}{{\mathbf{K}}_{2}}{{\left[ {{{\mathbf{\dot{q}}}}^{a}} \right]}^{\frac{{{r}_{1}}}{{{r}_{2}}}}}. \\ 
\end{aligned}
\end{equation}
By integrating both sides of \eqref{eq26} over the time interval $\left[ 0,t \right]$, we have: 
\begin{equation}
\label{eq27}
\begin{aligned}
   & \int\limits_{0}^{t}{{{{\mathbf{\dot{q}}}}^{a\top}}\pmb{\tau }_{ext}^{a}d\tau }= \int\limits_{0}^{t}{{{{\mathbf{\dot{q}}}}^{a\top}}\mathbf{S}\left( {{\mathbf{q}}^{a}},{{{\mathbf{\dot{q}}}}^{a}} \right){{\left[ {{{\mathbf{\dot{q}}}}^{a}} \right]}^{\frac{{{r}_{1}}}{{{r}_{2}}}}}}d\tau \\ 
 & \,\,\,\,\,\,\,\,\,\,\,\,\,\,\,\,\,\,\,\,\,\,\,\,\,\,\,+{{k}^{a}}\left( {{\mathbf{q}}^{a}}\left( t \right),{{{\mathbf{\dot{q}}}}^{a}}\left( t \right) \right)-{{k}^{a}}\left( {{\mathbf{q}}^{a}}\left( 0 \right),{{{\mathbf{\dot{q}}}}^{a}}\left( 0 \right) \right). 
\end{aligned}
\end{equation}
Obviously, if  ${{r}_{1}},\,{{r}_{2}}$  $\left( {{r}_{1}}<{{r}_{2}} \right)$ are positive intergers and ${{\mathbf{K}}_{2}}$  denotes positive-definite matrix, ${{k}^{a}}\left( {{\mathbf{q}}^{a}},{{{\mathbf{\dot{q}}}}^{a}} \right)>0$ and $\int\limits_{0}^{t}{{{{\mathbf{\dot{q}}}}^{a\top}}\mathbf{S}\left( {{\mathbf{q}}^{a}},{{{\mathbf{\dot{q}}}}^{a}} \right){{\left[ {{{\mathbf{\dot{q}}}}^{a}} \right]}^{\frac{{{r}_{1}}}{{{r}_{2}}}}}}d\tau>0$ under ${{k}^{a}}\left( {{\mathbf{q}}^{a}},{{{\mathbf{\dot{q}}}}^{a}} \right)-k_{d}^{a}\ge 0$. Therefore, the closed-loop system is strictly passive when the kinetic energy exceeds the predetermined level. \hspace{1em plus 1fill}\IEEEQEDhere

\end{proof1}

\begin{proof1}
\label{pt2}
According to Lyapunov's direct method in \cite{cL1,cL2}, a Lyapunov function is selected  for examining the convergence of  the kinetic energy as follows:  
\begin{equation}
\label{eq28}
{{V}_{1}}\left( {{\mathbf{q}}^{a}},{{{\mathbf{\dot{q}}}}^{a}} \right)=\frac{1}{2}{{\left( {{k}^{a}}\left( {{\mathbf{q}}^{a}},{{{\mathbf{\dot{q}}}}^{a}} \right)-k_{d}^{a} \right)}^{2}}.
\end{equation}
Taking the derivative of \eqref{eq28} with respect to time and combining it with the kinetic energy in \eqref{eq11}, we have:
\begin{equation}
\label{eq29}
\begin{aligned}
  & {{{\dot{V}}}_{1}}\left( {{\mathbf{q}}^{a}},{{{\mathbf{\dot{q}}}}^{a}} \right) =\left( {{k}^{a}}\left( {{\mathbf{q}}^{a}},{{{\mathbf{\dot{q}}}}^{a}} \right)-k_{d}^{a} \right)\left( {{{\dot{k}}}^{a}}\left( {{\mathbf{q}}^{a}},{{{\mathbf{\dot{q}}}}^{a}} \right)-\dot{k}_{d}^{a} \right) \\ 
 & =\left( {{k}^{a}}\left( {{\mathbf{q}}^{a}},{{{\mathbf{\dot{q}}}}^{a}} \right)-k_{d}^{a} \right)\left( {{{\mathbf{\dot{q}}}}^{a\top}}{{\mathbf{M}}^{a}}{{{\mathbf{\ddot{q}}}}^{a}}+\frac{1}{2}{{{\mathbf{\dot{q}}}}^{a\top}}{{{\mathbf{\dot{M}}}}^{a}}{{{\mathbf{\dot{q}}}}^{a}} \right). \\ 
\end{aligned}
\end{equation}
Substituting the augmented system \eqref{eq6} into \eqref{eq29}, yields:
\begin{equation}
\label{eq30}
{{\dot{V}}_{1}}=\left( {{k}^{a}}\left( {{\mathbf{q}}^{a}},{{{\mathbf{\dot{q}}}}^{a}} \right)-k_{d}^{a} \right)\left( \begin{aligned}
  & \,{{{\mathbf{\dot{q}}}}^{a\top}}\left( {{\pmb{\tau }}^{a}}+\pmb{\tau }_{ext}^{a}-{{\mathbf{C}}^{a}}{{{\mathbf{\dot{q}}}}^{a}} \right) \\ 
 & \,\,\,\,\,\,\,\,\,\,\,\,\,\,\,\,\,\,\,\,\,+\frac{1}{2}{{{\mathbf{\dot{q}}}}^{a\top}}{{{\mathbf{\dot{M}}}}^{a}}{{{\mathbf{\dot{q}}}}^{a}} \\ 
\end{aligned} \right)
\end{equation}
Considering $\pmb{\tau }_{ext}^{a}=\mathbf{0}$ and adopting the time-varying PVFC in \eqref{eq12}, we obtain,
\begin{equation}
\label{eq31}
{{\dot{V}}_{1}}=\left( {{k}^{a}}\left( {{\mathbf{q}}^{a}},{{{\mathbf{\dot{q}}}}^{a}} \right)-k_{d}^{a} \right)\left( \begin{aligned}
  & \frac{1}{2}{{{\mathbf{\dot{q}}}}^{a\top}}\left( {{{\mathbf{\dot{M}}}}^{a}}-2{{\mathbf{C}}^{a}} \right){{{\mathbf{\dot{q}}}}^{a}} \\ 
 & +{{{\mathbf{\dot{q}}}}^{a\top}}\mathbf{G}\left( {{\mathbf{q}}^{a}},{{{\mathbf{\dot{q}}}}^{a}},t \right){{{\mathbf{\dot{q}}}}^{a}} \\ 
 & +{{{\mathbf{\dot{q}}}}^{a\top}}\mathbf{R}\left( {{\mathbf{q}}^{a}},{{{\mathbf{\dot{q}}}}^{a}},t \right){{{\mathbf{\dot{q}}}}^{a}} \\ 
 & -{{{\mathbf{\dot{q}}}}^{a\top}}\mathbf{S}\left( {{\mathbf{q}}^{a}},{{{\mathbf{\dot{q}}}}^{a}} \right){{\left[ {{{\mathbf{\dot{q}}}}^{a}} \right]}^{\frac{{{r}_{1}}}{{{r}_{2}}}}} \\ 
\end{aligned} \right)
\end{equation}
It is straightforward to derive the following formulation using the properties of skew-symmetric matrices.
\begin{equation}
\label{eq32}
\begin{aligned}
  & {{{\dot{V}}}_{1}}=-\left( {{k}^{a}}\left( {{\mathbf{q}}^{a}},{{{\mathbf{\dot{q}}}}^{a}} \right)-k_{d}^{a} \right)s\left( e \right)\left( {{{\mathbf{\dot{q}}}}^{a\top}}{{\mathbf{K}}_{2}}{{\left[ {{{\mathbf{\dot{q}}}}^{a}} \right]}^{\frac{{{r}_{1}}}{{{r}_{2}}}}} \right) \\ 
\end{aligned}.
\end{equation}
Based on the saturation function \eqref{eq17}, \eqref{eq32} is evaluated as,
\begin{equation}
\label{eq33}
\begin{aligned}
  & {{{\dot{V}}}_{1}}\,\le -{{\lambda }_{\min }}\left( s\left( e \right) \right)\left| {{k}^{a}}\left( {{\mathbf{q}}^{a}},{{{\mathbf{\dot{q}}}}^{a}} \right)-k_{d}^{a} \right|L\left( {{{\mathbf{\dot{q}}}}^{a}} \right) \\ 
 & \,\,\,\,\,\,\,\,\le -{{\lambda }_{\min }}\left( s\left( e \right) \right)\sqrt{2{{V}_{1}}\left( {{\mathbf{q}}^{a}},{{{\mathbf{\dot{q}}}}^{a}} \right)}L\left( {{{\mathbf{\dot{q}}}}^{a}} \right), \\ 
\end{aligned}
\end{equation}
in which $L\left( {{{\mathbf{\dot{q}}}}^{a}} \right)={{\mathbf{\dot{q}}}^{a\top}}{{\mathbf{K}}_{2}}{{\left[ {{{\mathbf{\dot{q}}}}^{a}} \right]}^{\frac{{{r}_{1}}}{{{r}_{2}}}}}$ and ${{\lambda }_{\min }}\left( s\left( e \right) \right)=\min \left\{ \frac{\left| {{\eta }_{\min }} \right|}{2}\left( 1-\cos \left( \frac{\pi }{{{\delta }_{1}}}e \right) \right),\frac{{{\eta }_{\max }}}{2}\left( 1-\cos \left( \frac{\pi }{{{\delta }_{2}}}e \right) \right) \right\}$. Each term in  ${{\mathbf{M}}^{a}}\left( {{\mathbf{q}}^{a}} \right)$ and ${{\mathbf{K}}_{2}}$ is a trigonometric function of the angular position or constant. Moreover,  ${{\mathbf{M}}^{a}}\left( {{\mathbf{q}}^{a}} \right)$ is symmetric and positive definite. ${{\mathbf{K}}_{2}}$ is diagonal  and positive definite. Therefore, the above matrices satisfy the following inequality.
\begin{equation}
\label{eq34}
\left\{ \begin{aligned}
  & {{k}^{a}}\left( {{\mathbf{q}}^{a}},{{{\mathbf{\dot{q}}}}^{a}} \right)=\frac{1}{2}{{{\mathbf{\dot{q}}}}^{a\top}}{{\mathbf{M}}^{a}}{{{\mathbf{\dot{q}}}}^{a}}\le \frac{1}{2}{{\lambda }_{\max }}\left( {{\mathbf{M}}^{a}} \right)\left\| {{{\mathbf{\dot{q}}}}^{a}} \right\|_{2}^{2} \\ 
 & {{k}^{a}}\left( {{\mathbf{q}}^{a}},{{{\mathbf{\dot{q}}}}^{a}} \right)=\frac{1}{2}{{{\mathbf{\dot{q}}}}^{a\top}}{{\mathbf{M}}^{a}}{{{\mathbf{\dot{q}}}}^{a}}\ge \frac{1}{2}{{\lambda }_{\min }}\left( {{\mathbf{M}}^{a}} \right)\left\| {{{\mathbf{\dot{q}}}}^{a}} \right\|_{2}^{2}, \\ 
\end{aligned} \right.
\end{equation}
\begin{equation}
\label{eq35}
\left\{ \begin{aligned}
  & {{{\mathbf{\dot{q}}}}^{a\top}}{{\mathbf{K}}_{2}}{{\left[ {{{\mathbf{\dot{q}}}}^{a}} \right]}^{\frac{{{r}_{1}}}{{{r}_{2}}}}}\le {{\lambda }_{\max }}\left( {{\mathbf{K}}_{2}} \right){{\left( \left\| {{{\mathbf{\dot{q}}}}^{a}} \right\|_{\frac{{{r}_{1}}+{{r}_{2}}}{{{r}_{2}}}}^{{}} \right)}^{\frac{{{r}_{1}}+{{r}_{2}}}{{{r}_{2}}}}} \\ 
 & {{{\mathbf{\dot{q}}}}^{a\top}}{{\mathbf{K}}_{2}}{{\left[ {{{\mathbf{\dot{q}}}}^{a}} \right]}^{\frac{{{r}_{1}}}{{{r}_{2}}}}}\ge {{\lambda }_{\min }}\left( {{\mathbf{K}}_{2}} \right){{\left( \left\| {{{\mathbf{\dot{q}}}}^{a}} \right\|_{\frac{{{r}_{1}}+{{r}_{2}}}{{{r}_{2}}}}^{{}} \right)}^{\frac{{{r}_{1}}+{{r}_{2}}}{{{r}_{2}}}}} \\ 
\end{aligned} \right.
\end{equation}
where ${{\lambda }_{\min }}\left( \bullet  \right)$ and ${{\lambda }_{\max }}\left( \bullet  \right)$ are the smallest and largest eigenvalues of the matrix $\left( \bullet  \right)$. Selecting the positive integers ${{r}_{1}},\,{{r}_{2}}$ $\left( {{r}_{1}}<{{r}_{2}} \right)$ and applying \eqref{eq35}, Lemma \ref{lm1}, and \eqref{eq34} for $L\left( {{{\mathbf{\dot{q}}}}^{a}} \right)$, respectively, yields,
\begin{equation}
\label{eq36}
\begin{aligned}
  & L\left( {{{\mathbf{\dot{q}}}}^{a}} \right)\,\ge {{\lambda }_{\min }}\left( {{\mathbf{K}}_{2}} \right){{\left( \left\| {{{\mathbf{\dot{q}}}}^{a}} \right\|_{\frac{{{r}_{1}}+{{r}_{2}}}{{{r}_{2}}}}^{{}} \right)}^{\frac{{{r}_{1}}+{{r}_{2}}}{{{r}_{2}}}}} \\ 
 & \,\,\,\,\,\,\,\,\,\,\,\,\,\,\,\,\ge {{\lambda }_{\min }}\left( {{\mathbf{K}}_{2}} \right){{\left( \left\| {{{\mathbf{\dot{q}}}}^{a}} \right\|_{2}^{2} \right)}^{\frac{{{r}_{1}}+{{r}_{2}}}{2{{r}_{2}}}}} \\ 
 & \,\,\,\,\,\,\,\,\,\,\,\,\,\,\,\ge {{\lambda }_{\min }}\left( {{\mathbf{K}}_{2}} \right){{\left( \frac{2{{k}^{a}}\left( {{\mathbf{q}}^{a}},{{{\mathbf{\dot{q}}}}^{a}} \right)}{{{\lambda }_{\max }}\left( {{\mathbf{M}}^{a}} \right)} \right)}^{\frac{{{r}_{1}}+{{r}_{2}}}{2{{r}_{2}}}}} \\ 
\end{aligned}
\end{equation}
Substituting \eqref{eq36} into \eqref{eq33}, we obtain, 
\begin{equation}
\label{eq37}
{{\dot{V}}_{1}}\le -{{\lambda }_{\min }}\left( s\left( e \right) \right){{\lambda }_{\min }}\left( {{\mathbf{K}}_{2}} \right)\sqrt{2{{V}_{1}}}{{\left( \frac{2{{k}^{a}}\left( {{\mathbf{q}}^{a}},{{{\mathbf{\dot{q}}}}^{a}} \right)}{{{\lambda }_{\max }}\left( {{\mathbf{M}}^{a}} \right)} \right)}^{\frac{{{r}_{1}}+{{r}_{2}}}{2{{r}_{2}}}}}
\end{equation}
Thus, by designing appropriate parameters, ${{V}_{1}}\left( {{\mathbf{q}}^{a}},{{{\mathbf{\dot{q}}}}^{a}} \right)\,$ is positive-definite, ${{\dot{V}}_{1}}\left( {{\mathbf{q}}^{a}},{{{\mathbf{\dot{q}}}}^{a}} \right)\,$ is negative definite, and ${{\dot{V}}_{1}}\left( {{\mathbf{q}}^{a}},{{{\mathbf{\dot{q}}}}^{a}} \right)\,=0$ if and only if ${{k}^{a}}\left( {{\mathbf{q}}^{a}},{{{\mathbf{\dot{q}}}}^{a}} \right)-k_{d}^{a}=0$. To compute the convergence time of the kinetic energy, the following two scenarios are considered.

For ${{k}^{a}}\left( {{\mathbf{q}}^{a}}, {{{\mathbf{\dot{q}}}}^{a}} \right)\ge k_{d}^{a}$, combining \eqref{eq28} with \eqref{eq37}, yields:
\begin{equation}
\label{eq38}
{{\dot{V}}_{1}}\,\,\le -\psi \sqrt{2{{V}_{1}}}{{\left( \sqrt{2{{V}_{1}}}+k_{d}^{a} \right)}^{\frac{{{r}_{1}}+{{r}_{2}}}{2{{r}_{2}}}}}
\end{equation}
with $\psi =\frac{{{2}^{\frac{{{r}_{1}}+{{r}_{2}}}{2{{r}_{2}}}}}{{\lambda }_{\min }}\left( s\left( e \right) \right){{\lambda }_{\min }}\left( {{\mathbf{K}}_{2}} \right)}{{{\left( {{\lambda }_{\max }}\left( {{\mathbf{M}}^{a}} \right) \right)}^{\frac{{{r}_{1}}+{{r}_{2}}}{2{{r}_{2}}}}}}$. We rewrite \eqref{eq38} in the following form.
\begin{equation}
\label{eq39}
\begin{aligned}
  & d{{t}}\,\,\le -\frac{d{{V}_{1}}}{\psi \sqrt{2{{V}_{1}}}{{\left( \sqrt{2{{V}_{1}}}+k_{d}^{a} \right)}^{\frac{{{r}_{1}}+{{r}_{2}}}{2{{r}_{2}}}}}} \\ 
 & \,\,\,\,\,\,\,\,\le -\frac{d\left( \sqrt{2{{V}_{1}}}+k_{d}^{a} \right)}{\psi {{\left( \sqrt{2{{V}_{1}}}+k_{d}^{a} \right)}^{\frac{{{r}_{1}}+{{r}_{2}}}{2{{r}_{2}}}}}} \\ 
\end{aligned}
\end{equation}
Taking the integration \eqref{eq39}, we obtain:
\begin{equation}
\label{eq40}
\int\limits_{0}^{{{T}_{1}}}{dt}\,\,\le -\int\limits_{{{V}_{1}}\left( 0 \right)}^{0}{\frac{d\left( \sqrt{2{{V}_{1}}}+k_{d}^{a} \right)}{\psi {{\left( \sqrt{2{{V}_{1}}}+k_{d}^{a} \right)}^{\frac{{{r}_{1}}+{{r}_{2}}}{2{{r}_{2}}}}}}}
\end{equation}
Calculating \eqref{eq40}, the convergence time of the first scenario is presented as,
\begin{equation}
\label{eq41}
{{T}_{1}}\le \frac{2{{r}_{2}}}{\psi \left( {{r}_{2}}-{{r}_{1}} \right)}\left( {{\left( \sqrt{2{{V}_{1}}\left( 0 \right)}+k_{d}^{a} \right)}^{\frac{{{r}_{2}}-{{r}_{1}}}{2{{r}_{2}}}}}-{{\left( k_{d}^{a} \right)}^{\frac{{{r}_{2}}-{{r}_{1}}}{2{{r}_{2}}}}} \right).
\end{equation}

For ${{k}^{a}}\left( {{\mathbf{q}}^{a}},{{{\mathbf{\dot{q}}}}^{a}} \right)<k_{d}^{a}$, substituting \eqref{eq28} into \eqref{eq37}, the second scenario is given as follows:
\begin{equation}
\label{eq42}
{{\dot{V}}_{1}}\,\,\le -\psi \sqrt{2{{V}_{1}}}{{\left( k_{d}^{a}-\sqrt{2{{V}_{1}}} \right)}^{\frac{{{r}_{1}}+{{r}_{2}}}{2{{r}_{2}}}}}
\end{equation}
According to the same approach in \eqref{eq39} and \eqref{eq40}, the convergence time for the second scenario is computed as follows.
\begin{equation}
\label{eq43}
{{T}_{2}}\le \frac{2{{r}_{2}}}{\psi \left( {{r}_{2}}-{{r}_{1}} \right)}\left( {{\left( k_{d}^{a} \right)}^{\frac{{{r}_{2}}-{{r}_{1}}}{2{{r}_{2}}}}}-{{\left( k_{d}^{a}-\sqrt{2{{V}_{1}}\left( 0 \right)} \right)}^{\frac{{{r}_{2}}-{{r}_{1}}}{2{{r}_{2}}}}} \right).
\end{equation}
Thus, the kinetic energy of the augmented system converges to the predetermined level within the finite time interval in \eqref{eq41} and \eqref{eq43}. \hspace{1em plus 1fill}\IEEEQEDhere
\end{proof1}

In order to prove Theorem \ref{thm3}, two supplementary lemmas are utilized.
\begin{lemma}
\label{lm4}
According to the $\alpha \left( t \right)$ term in \eqref{eq21} and settling time in \eqref{eq19}, the differential of $\alpha \left( t \right)$ holds in the absence of external disturbance  $\left( \pmb{\tau }_{ext}^{a}=0 \right)$.  
\begin{equation}
\label{eq44.1}
\underset{t\to T}{\mathop{\lim }}\,\dot{\alpha }\left( t \right)=0.
\end{equation}

\renewcommand{\proof2}{\emph{Proof of Lemma \ref{lm4}:}}
By calculating the limitation of \eqref{eq21} over the settling time interval $\left[ 0,T \right]$, we obtain:
\begin{equation}
\label{eq44.2}
\underset{t\to T}{\mathop{\lim }}\,\alpha \left( t \right)=\underset{t\to T}{\mathop{\lim }}\,\sqrt{\frac{{{k}^{a}}\left( {{\mathbf{q}}^{a}}\left( t \right),{{{\mathbf{\dot{q}}}}^{a}}\left( t \right) \right)}{{{E}^{a}}}}=\sqrt{\frac{k_{d}^{a}}{{{E}^{a}}}}.
\end{equation}
Because the limit in \eqref{eq44.2} is constant; therefore, $\dot{\alpha }\left( t \right)$ converges to zero if only $t\to T$. \hspace{1em plus 1fill}\IEEEQEDhere
\end{lemma}

\begin{lemma}
\label{lm5}
The following formulations hold:  
\begin{equation}
\label{eq44}
{{\mathbf{w}}^{\top}}{{\mathbf{V}}^{a}}\left( \mathbf{q},t \right)=0,
\end{equation}
\begin{equation}
\label{eq45}
\mathbf{G}\left( {{\mathbf{q}}^{a}},{{{\mathbf{\dot{q}}}}^{a}},t \right){{\mathbf{\dot{q}}}^{a}}-\alpha\left(t\right) \mathbf{w}=\mathbf{G}\left( {{\mathbf{q}}^{a}},{{{\mathbf{\dot{q}}}}^{a}},t \right){{\mathbf{e}}_{\mathbf{v}}}.
\end{equation}

\renewcommand{\proof2}{\emph{Proof of Lemma \ref{lm5}:}}
Inspired by the time-invariant PVFC \cite{c18}, Lemma \ref{lm5} is extended for our time-varying PVFC. As a result, the expressions in \eqref{eq44} and \eqref{eq45} are satisfied. \hspace{1em plus 1fill}\IEEEQEDhere
\end{lemma}
 
\begin{proof1}
\label{pt3}
The following Lyapunov function is selected to analyze the system stability using Lyapunov's direct method in \cite{cL1,cL2}.
\begin{equation}
\label{eq46}
{{V}_{2}}\left( {{\mathbf{q}}^{a}},{{{\mathbf{\dot{q}}}}^{a}} \right)=\frac{1}{2}\mathbf{e}_{\mathbf{q}}^{\top}\mathbf{e}_{\mathbf{q}}^{{}}+\frac{1}{2}\mathbf{e}_{\mathbf{v}}^{\top}{{\mathbf{M}}^{a}}\mathbf{e}_{\mathbf{v}}^{{}}>0.
\end{equation}
Differentiating \eqref{eq46} and adopting the augmented system in \eqref{eq6} and Lemma \ref{lm4}, we have: 
\begin{equation}
\label{eq47}
\begin{aligned}
  & {{{\dot{V}}}_{2}}\left( {{\mathbf{q}}^{a}},{{{\mathbf{\dot{q}}}}^{a}} \right)=\mathbf{e}_{\mathbf{q}}^{\top}\mathbf{\dot{e}}_{\mathbf{q}}^{{}}+\mathbf{e}_{\mathbf{v}}^{\top}{{\mathbf{M}}^{a}}\mathbf{\dot{e}}_{\mathbf{v}}^{{}}+\frac{1}{2}\mathbf{e}_{\mathbf{v}}^{\top}{{{\mathbf{\dot{M}}}}^{a}}\mathbf{e}_{\mathbf{v}}^{{}} \\ 
 & \,\,\,\,\,\,\,\,\,\,\,\,\,\,\,\,\,\,\,\,\,\,\,\,\,\,\,=\frac{1}{2}\mathbf{e}_{\mathbf{v}}^{\top}{{{\mathbf{\dot{M}}}}^{a}}\mathbf{e}_{\mathbf{v}}^{{}}+\mathbf{e}_{\mathbf{q}}^{\top}\left( \mathbf{\dot{\mathbf{q}}}-\mathbf{\dot{Q}} \right) \\ 
 & \,\,\,\,\,\,\,\,\,\,\,\,\,\,\,\,\,\,\,\,\,\,\,\,\,\,\,\,\,\,\,\,\,\,+\mathbf{e}_{\mathbf{v}}^{\top}\left( {{\pmb{\tau }}^{a}}+\pmb{\tau }_{ext}^{a}-{{\mathbf{C}}^{a}}\left( {{\mathbf{q}}^{a}},{{{\mathbf{\dot{q}}}}^{a}} \right){{{\mathbf{\dot{q}}}}^{a}} \right) \\ 
 & \,\,\,\,\,\,\,\,\,\,\,\,\,\,\,\,\,\,\,\,\,\,\,\,\,\,\,\,\,\,\,\,\,\,\,\,\,\,\,\,\,\,\,\,\,\,\,\,\,\,\,\,\,\,\,\,\,\,\,\,-\alpha \left( t \right)\mathbf{e}_{\mathbf{v}}^{\top}{{\mathbf{M}}^{a}}{{{\mathbf{\dot{V}}}}^{a}}\left( {{\mathbf{q}}^{a}},t \right) \\ 
\end{aligned}
\end{equation}
Substituting the time-varying velocity field in \eqref{eq8} and the improved PVFC in \eqref{eq12} into \eqref{eq47} in absence of external force, yields: 
\begin{equation}
\label{eq48}
\begin{aligned}
  & {{{\dot{V}}}_{2}}=\frac{1}{2}\mathbf{e}_{\mathbf{v}}^{\top}{{{\mathbf{\dot{M}}}}^{a}}\mathbf{e}_{\mathbf{v}}^{{}}+\mathbf{e}_{\mathbf{q}}^{\top}\left( \mathbf{\dot{Q}}-{{\mathbf{K}}_{1}}\left( \mathbf{q}-\mathbf{Q} \right)-\mathbf{\dot{Q}} \right) \\ 
 & \,\,\,\,\,\,\,\,\,\,+\mathbf{e}_{\mathbf{v}}^{\top}\left( \mathbf{G}\left( {{\mathbf{q}}^{a}},{{{\mathbf{\dot{q}}}}^{a}},t \right){{{\mathbf{\dot{q}}}}^{a}}+\mathbf{R}\left( {{\mathbf{q}}^{a}},{{{\mathbf{\dot{q}}}}^{a}},t \right){{{\mathbf{\dot{q}}}}^{a}} \right) \\ 
 & \,\,\,\,\,\,\,\,\,\,-\mathbf{e}_{\mathbf{v}}^{\top}{{\mathbf{C}}^{a}}\left( {{\mathbf{q}}^{a}},{{{\mathbf{\dot{q}}}}^{a}} \right)\left( \mathbf{e}_{\mathbf{v}}^{{}}+\alpha \left( t \right){{\mathbf{V}}^{a}}\left( {{\mathbf{q}}^{a}},t \right) \right) \\ 
 & \,\,\,\,\,\,\,\,\,\,-\mathbf{e}_{\mathbf{v}}^{\top}\mathbf{S}\left( {{\mathbf{q}}^{a}},{{{\mathbf{\dot{q}}}}^{a}} \right){{\left[ {{{\mathbf{\dot{q}}}}^{a}} \right]}^{\frac{{{r}_{1}}}{{{r}_{2}}}}}-\alpha \left( t \right)\mathbf{e}_{\mathbf{v}}^{\top}{{\mathbf{M}}^{a}}{{{\mathbf{\dot{V}}}}^{a}}\left( {{\mathbf{q}}^{a}},t \right) \\ 
\end{aligned}
\end{equation}
We rewrite \eqref{eq48} by using the definition in \eqref{eq14}. 
\begin{equation}
\label{eq49}
\begin{aligned}
  & {{{\dot{V}}}_{2}}\,=\frac{1}{2}\mathbf{e}_{\mathbf{v}}^{\top}\left( {{{\mathbf{\dot{M}}}}^{a}}-2{{\mathbf{C}}^{a}}\left( {{\mathbf{q}}^{a}},{{{\mathbf{\dot{q}}}}^{a}} \right) \right)\mathbf{e}_{\mathbf{v}}^{{}}-\mathbf{e}_{\mathbf{q}}^{\top}{{\mathbf{K}}_{1}}\mathbf{e}_{\mathbf{q}}^{{}} \\ 
 & \,\,\,\,\,\,\,\,+\mathbf{e}_{\mathbf{v}}^{\top}\mathbf{R}\left( {{\mathbf{q}}^{a}},{{{\mathbf{\dot{q}}}}^{a}},t \right){{{\mathbf{\dot{q}}}}^{a}}-\mathbf{e}_{\mathbf{v}}^{\top}\mathbf{S}\left( {{\mathbf{q}}^{a}},{{{\mathbf{\dot{q}}}}^{a}} \right){{\left[ {{{\mathbf{\dot{q}}}}^{a}} \right]}^{\frac{{{r}_{1}}}{{{r}_{2}}}}} \\ 
 & \,\,\,\,\,\,\,\,+\mathbf{e}_{\mathbf{v}}^{\top}\left( \mathbf{G}\left( {{\mathbf{q}}^{a}},{{{\mathbf{\dot{q}}}}^{a}},t \right){{{\mathbf{\dot{q}}}}^{a}}-\alpha \left( t \right)\mathbf{w} \right) \\ 
\end{aligned}
\end{equation}
Applying Lemma \ref{lm5}, we obtain:
\begin{equation}
\label{eq50}
\begin{aligned}
  & {{{\dot{V}}}_{2}}=\frac{1}{2}\mathbf{e}_{\mathbf{v}}^{\top}\left( {{{\mathbf{\dot{M}}}}^{a}}-2{{\mathbf{C}}^{a}}\left( {{\mathbf{q}}^{a}},{{{\mathbf{\dot{q}}}}^{a}} \right) \right)\mathbf{e}_{\mathbf{v}}^{{}}-\mathbf{e}_{\mathbf{q}}^{\top}{{\mathbf{K}}_{1}}\mathbf{e}_{\mathbf{q}}^{{}} \\ 
 & \,\,\,\,\,\,\,\,+ {{{\mathbf{\dot{q}}}}^{a\top}} \mathbf{R}\left( {{\mathbf{q}}^{a}},{{{\mathbf{\dot{q}}}}^{a}},t \right){{{\mathbf{\dot{q}}}}^{a}} \\
 & \,\,\,\,\,\,\,\,-\alpha \left( t \right){{\mathbf{V}}^{a\top}}\left( {{\mathbf{q}}^{a}},t \right)\mathbf{R}\left( {{\mathbf{q}}^{a}},{{{\mathbf{\dot{q}}}}^{a}},t \right){{{\mathbf{\dot{q}}}}^{a}}\, \\ 
 & \,\,\,\,\,\,\,\,+\mathbf{e}_{\mathbf{v}}^{\top}\mathbf{G}\left( {{\mathbf{q}}^{a}},{{{\mathbf{\dot{q}}}}^{a}},t \right){{\mathbf{e}}_{\mathbf{v}}}-\mathbf{e}_{\mathbf{v}}^{\top}\mathbf{S}\left( {{\mathbf{q}}^{a}},{{{\mathbf{\dot{q}}}}^{a}} \right){{\left[ {{{\mathbf{\dot{q}}}}^{a}} \right]}^{\frac{{{r}_{1}}}{{{r}_{2}}}}} \\ 
\end{aligned}
\end{equation}
As ${{\mathbf{\dot{M}}}^{a}}-2{{\mathbf{C}}^{a}}\left( {{\mathbf{q}}^{a}},{{{\mathbf{\dot{q}}}}^{a}} \right),\,\,\mathbf{G}\left( {{\mathbf{q}}^{a}},{{{\mathbf{\dot{q}}}}^{a}},t \right),\,\,\mathbf{R}\left( {{\mathbf{q}}^{a}},{{{\mathbf{\dot{q}}}}^{a}},t \right)$ are skew-symmetric matrices; therefore, equation \eqref{eq50} is presented in the following form.
\begin{equation}
\label{eq51}
\begin{aligned}
  & {{{\dot{V}}}_{2}}=-\mathbf{e}_{\mathbf{q}}^{\top}{{\mathbf{K}}_{1}}\mathbf{e}_{\mathbf{q}}^{{}}-\alpha \left( t \right){{\mathbf{V}}^{a\top}}\left( {{\mathbf{q}}^{a}},t \right)\mathbf{R}\left( {{\mathbf{q}}^{a}},{{{\mathbf{\dot{q}}}}^{a}},t \right){{{\mathbf{\dot{q}}}}^{a}}\, \\ 
 & \,\,\,\,\,\,\,\,\,\,\,\,\,\,\,\,\,\,\,\,\,\,\,\,\,\,\,\,\,\,\,\,\,\,\,\,\,\,\,\,\,\,\,\,\,\,\,\,\,\,\,\,\,\,\,\,\,\,\,\,\,\,\,\,\,\,\,\,\,\,-\mathbf{e}_{\mathbf{v}}^{\top}\mathbf{S}\left( {{\mathbf{q}}^{a}},{{{\mathbf{\dot{q}}}}^{a}} \right){{\left[ {{{\mathbf{\dot{q}}}}^{a}} \right]}^{\frac{{{r}_{1}}}{{{r}_{2}}}}} \\ 
\end{aligned}
\end{equation}
After the settling time in \eqref{eq19}, we substitute $\mathbf{R}\left( {{\mathbf{q}}^{a}},{{{\mathbf{\dot{q}}}}^{a}},t \right)$ in \eqref{eq13} into \eqref{eq51}. 
\begin{equation}
\label{eq52}
\begin{aligned}
  & {{{\dot{V}}}_{2}}=\,-\mathbf{e}_{\mathbf{q}}^{\top}{{\mathbf{K}}_{1}}\mathbf{e}_{\mathbf{q}}^{{}} \\ 
 & -\alpha \left( t \right)\kappa \left( \begin{aligned}
  & {{\mathbf{V}}^{a\top}}\left( {{\mathbf{q}}^{a}},t \right){{\mathbf{M}}^{a}}\left( {{\mathbf{q}}^{a}} \right){{\mathbf{V}}^{a}}\left( {{\mathbf{q}}^{a}},t \right){{{\mathbf{\dot{q}}}}^{a\top}}{{\mathbf{M}}^{a}}\left( {{\mathbf{q}}^{a}} \right){{{\mathbf{\dot{q}}}}^{a}} \\ 
 & \,\,\,\,\,\,\,\,\,\,\,\,\,\,-{{\left( {{\mathbf{V}}^{a\top}}\left( {{\mathbf{q}}^{a}},t \right){{\mathbf{M}}^{a}}\left( {{\mathbf{q}}^{a}} \right){{{\mathbf{\dot{q}}}}^{a}} \right)}^{2}} \\ 
\end{aligned} \right) \\ 
\end{aligned}
\end{equation}
Adopting the Cauchy-Schwarz inequality in \cite{c40}, the following formulation holds. 
\begin{equation}
\left( \begin{aligned}
  & {{\mathbf{V}}^{a\top}}\left( {{\mathbf{q}}^{a}},t \right){{\mathbf{M}}^{a}}\left( {{\mathbf{q}}^{a}} \right){{\mathbf{V}}^{a}}\left( {{\mathbf{q}}^{a}},t \right){{{\mathbf{\dot{q}}}}^{a\top}}{{\mathbf{M}}^{a}}\left( {{\mathbf{q}}^{a}} \right){{{\mathbf{\dot{q}}}}^{a}} \\ 
 & \,\,\,\,\,\,\,\,\,\,\,\,\,\,\,-{{\left( {{\mathbf{V}}^{a\top}}\left( {{\mathbf{q}}^{a}},t \right){{\mathbf{M}}^{a}}\left( {{\mathbf{q}}^{a}} \right){{{\mathbf{\dot{q}}}}^{a}} \right)}^{2}} \\ 
\end{aligned} \right)\ge 0.
\label{eq53}
\end{equation}
Thus, by designing the positive-definite control parameters $\left( {{\mathbf{K}}_{1}},\,\kappa  \right)$ and incorporating the conditions in  \eqref{eq21} and \eqref{eq53}, ${{\mathbf{e}}_{\mathbf{q}}}=\mathbf{0}$ and ${{\mathbf{e}}_{\mathbf{v}}}=\mathbf{0}$ to the system is Lyapunov stable. \hspace{1em plus 1fill}\IEEEQEDhere
\end{proof1}

\begin{remark}
\label{r2}
According to the Proofs of Theorems \ref{pt1} and \ref{pt2}, selecting the control parameters such that ${{\eta }_{\max }}>0$, ${{\delta}_{2}}>0$, $r_1>0$, $r_2>0$, ${{\eta }_{\min }}<0$, ${{\delta }_{1}}<0$, and ${{\mathbf{K}}_{2}}$ is a positive-definite matrix ensures the convergence of the energy level. From \eqref{eq41} and \eqref{eq43}, the condition ${r_1}<{r_2}$ must hold to satisfy the feasibility of the convergence time. Furthermore, the positive-definite control parameters $\left( {{\mathbf{K}}_{1}},\,\kappa  \right)$ are also designed to achieve the stability of the closed-loop robotic system as presented in Proof of Theorem \ref{pt3}. However, configuring excessively large values of the control parameters may induce overshoot or actuator saturation, whereas excessively small values can degrade task performance and slow the convergence rate, as indicated in \eqref{eq41} and \eqref{eq43}. Thus, the control parameters should be fine-tuned to meet specific requirements such as task performance, convergence time, and actuator constraints, while fulfilling the aforementioned conditions. 
\end{remark}

\renewcommand{\proof2}{\emph{Proof of Corollary \ref{lm3}:}}
Due to the external force applied solely by the human operator, the power flow in co-carrying tasks is computed as follows:
\begin{equation}
{{P}_{r2h}}=-{{\mathbf{\dot{q}}}^{a\top}}\pmb{\tau }_{ext}^{a}.
\label{eq54}
\end{equation}
By substituting the augmented system in \eqref{eq6} and the improved PVFC in \eqref{eq12} into \eqref{eq54}, we obtain:
\begin{equation}
\begin{aligned}
  & {{P}_{r2h}}=\frac{1}{2}{{{\mathbf{\dot{q}}}}^{a\top}}\left( {{\mathbf{M}}^{a}}-2{{\mathbf{C}}^{a}}\left( {{\mathbf{q}}^{a}},{{{\mathbf{\dot{q}}}}^{a}} \right) \right){{{\mathbf{\dot{q}}}}^{a}} \\ 
 & \,\,\,\,\,\,\,\,\,\,\,\,\,\,\,\,\,\,\,+{{{\mathbf{\dot{q}}}}^{a\top}}\left( \mathbf{G}\left( {{\mathbf{q}}^{a}},{{{\mathbf{\dot{q}}}}^{a}},t \right)+\mathbf{R}\left( {{\mathbf{q}}^{a}},{{{\mathbf{\dot{q}}}}^{a}},t \right) \right){{{\mathbf{\dot{q}}}}^{a}} \\ 
 & \,\,\,\,\,\,\,\,\,\,\,\,\,\,\,\,\,\,\,-{{{\mathbf{\dot{q}}}}^{a\top}}\mathbf{S}\left( {{\mathbf{q}}^{a}},{{{\mathbf{\dot{q}}}}^{a}} \right){{\left[ {{{\mathbf{\dot{q}}}}^{a}} \right]}^{\frac{{{r}_{1}}}{{{r}_{2}}}}}-{\frac{d}{dt}\left( \frac{1}{2}{{{\mathbf{\dot{q}}}}^{aT}}{{\mathbf{M}}^{a}}{{{\mathbf{\dot{q}}}}^{a}} \right)} \\ 
\end{aligned}
\label{eq55}
\end{equation}
Adopting the property of skew-symmetric matrices and combining it with the definition in \eqref{eq11} and \eqref{eq13}, yields, 
\begin{equation}
\begin{aligned}
  & {{P}_{r2h}}=-{{{\mathbf{\dot{q}}}}^{a\top}}\mathbf{S}\left( {{\mathbf{q}}^{a}},{{{\mathbf{\dot{q}}}}^{a}} \right){{\left[ {{{\mathbf{\dot{q}}}}^{a}} \right]}^{\frac{{{r}_{1}}}{{{r}_{2}}}}}-\frac{d}{dt}\left( \frac{1}{2}{{{\mathbf{\dot{q}}}}^{a\top}}{{\mathbf{M}}^{a}}{{{\mathbf{\dot{q}}}}^{a}} \right) \\ 
 & \,\,\,\,\,\,\,\,\,\,\,=-s\left( {{k}^{a}}\left( \mathbf{q},{{\mathbf{q}}^{a}} \right)-k_{d}^{a} \right){{{\mathbf{\dot{q}}}}^{a\top}}{{\mathbf{K}}_{2}}{{\left[ {{{\mathbf{\dot{q}}}}^{a}} \right]}^{\frac{{{r}_{1}}}{{{r}_{2}}}}}-{{{\dot{k}}}^{a}} \\ 
\end{aligned}
\label{eq56}
\end{equation}

According to the energy-compensated control \eqref{eq12}, the energy of the augmented system exhibits a decreasing trend to approach $k_{d}^{a}$ in the case of ${{k}^{a}}\left( {{\mathbf{q}}^{a}}, {{{\mathbf{\dot{q}}}}^{a}} \right)\ge k_{d}^{a}$, resulting in the negative energy derivative ${{{\dot{k}}}^{a}}$. Furthermore, the positive upper bound of $s\left( {{k}^{a}}\left( \mathbf{q},{{\mathbf{q}}^{a}} \right)-k_{d}^{a} \right){{{\mathbf{\dot{q}}}}^{a\top}}{{\mathbf{K}}_{2}}{{\left[ {{{\mathbf{\dot{q}}}}^{a}} \right]}^{\frac{{{r}_{1}}}{{{r}_{2}}}}}$ can regulated by selecting the parameters ${{\mathbf{K}}_{2}},{{r}_{1}},r_{2}$ of \eqref{eq13} and of ${{\eta }_{\max }},{{\delta }_{2}}$
 \eqref{eq17} when ${{k}^{a}}\left( {{\mathbf{q}}^{a}},{{{\mathbf{\dot{q}}}}^{a}} \right)\ge k_{d}^{a}$. Meanwhile, for ${{k}^{a}}\left( {{\mathbf{q}}^{a}},{{{\mathbf{\dot{q}}}}^{a}} \right)< k_{d}^{a}$, the differential of energy is positive due to compensated action from \eqref{eq12} and the negative lower bound for the first right-hand side term in \eqref{eq56} can be decided by $ {{\mathbf{K}}_{2}},{{r}_{1}},r_{2}^{{}}, {{\eta }_{\max }},{{\delta }_{1}}$ in \eqref{eq13} and \eqref{eq17}. Thus, based on the aforementioned analyses, the power flow within the collaboration systems can be regulated by fine-tuning the control parameters $ {{\mathbf{K}}_{2}},{{r}_{1}},r_{2}^{{}}, {{\eta }_{\max }},{{\eta }_{\min }},{{\delta }_{1}},{{\delta }_{2}}$. Moreover, based on \eqref{eq54}-\eqref{eq56}, we also conclude that the power flow is zero under harmonious movement states with $\pmb{\tau }_{ext}^{a}=\mathbf{0}$. \hspace{1em plus 1fill}\IEEEQEDhere

\section{EXPERIMENTS}
To validate the efficiency of the proposed framework for human--robot co-carrying tasks, human-in-the-loop experiments are implemented in this section.

\subsection{Experimental Setup}

\begin{figure*}[!t]
\centering
\includegraphics[scale=0.65]{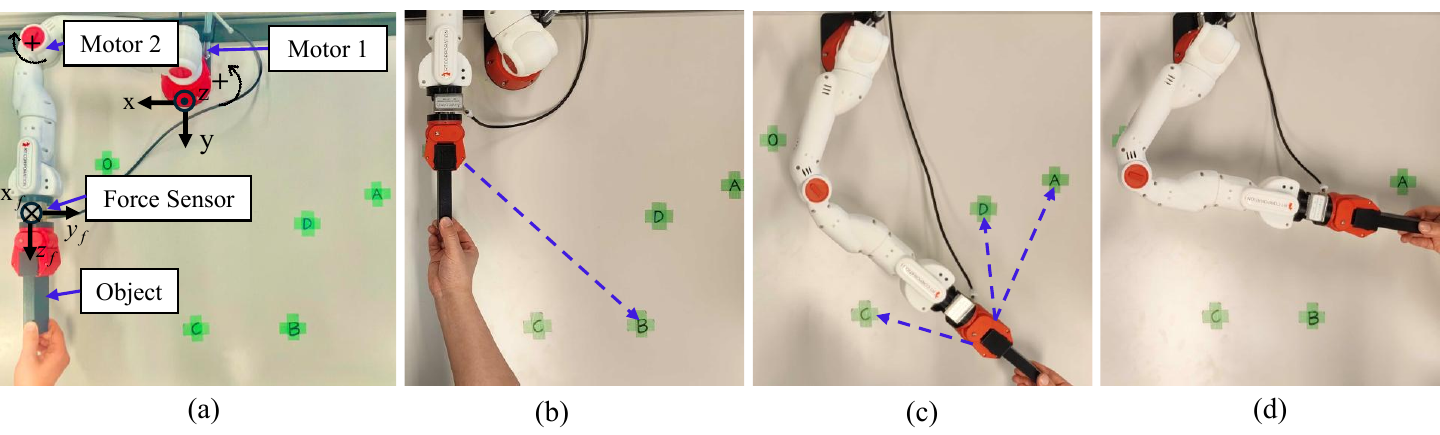}
\caption{The experimental framework of the human--robot co-carrying task, including (a) the experimental setup where a rigid object is collaboratively grasped by a human and a robot; (b) the object at its initial position (O) intended to toward position (B); (c) the object at position (B) transported to one of the target positions (A, C, or D); and (d) the object at the parking location (D).}
\label{f5}
\end{figure*}
\begin{table}[!t]
\caption{Parameters of control methods}
\label{table1}
\centering
\begin{tabular}{c|ccc}
\hline
Methods & Parameters \\
\hline
Proposed Method & ${{\mathbf{M}}_{d}}=10{{\mathbf{I}}_{2\times 2}}$, ${{\mathbf{D}}_{d}}=20{{\mathbf{I}}_{2\times 2}}$, \\ 
& ${{E}^{a}}=1000\,\text{J}$, $k_{a}^{d}=8\,\text{J}$, ${{\mathbf{K}}_{1}}=150{{\mathbf{I}}_{2\times 2}}$, \\
& ${{\mathbf{K}}_{2}}={{\mathbf{I}}_{3\times 3}}$, ${{r}_{1}}=9$, ${{r}_{2}}=11$,  \\
&  $\kappa =0.01$, ${{\delta }_{1}}=-0.1$, ${{\delta }_{2}}=0.1$,  \\
&
${{\eta }_{\min }}=-1$, ${{\eta }_{\max }}=1$.\\
\hline
Compared Methods 1 and 2 & ${{\mathbf{M}}_{d}}=10{{\mathbf{I}}_{2\times 2}}$, ${{\mathbf{D}}_{d}}=20{{\mathbf{I}}_{2\times 2}}$, \\ 
& ${{\mathbf{K}}_{P}}=3{{\mathbf{I}}_{2\times 2}}$, ${{\mathbf{K}}_{I}}=0.01{{\mathbf{I}}_{2\times 2}}$,  \\
&
${{\mathbf{K}}_{D}}=0.35{{\mathbf{I}}_{2\times 2}}$.\\
\hline
\end{tabular}
\end{table} 

\begin{table}[!t]
\caption{Parameters of the deep LSTM model training}
\label{table2}
\centering
\begin{tabular}{c|ccc}
\hline
Parameters & Values \\
\hline
Number of LSTM layers & 2\\
\hline
Number of input vectors $\left(m\right)$ & 10\\
\hline
Hidden layer unit dimension & 256\\
\hline
Number of iterations & 250\\
\hline
Initial learning rate & 0.001\\
\hline
Training sample batch & 16\\
\hline
Learn rate drop period & 50\\
\hline
Learn rate drop factor & 0.5\\
\hline
Validation split & 0.2\\
\hline
Optimizer & Adam\\
\hline
Loss function & Mean square error\\
\hline
\end{tabular}
\end{table} 
The co-carrying experimental framework on the OXY surface is designed by utilizing a robot manipulator (CRANE-X7, RT Corporation),  a rigid co-transported object, and a force sensor (SFS034YA301U6, Leptrino) as shown in Fig. \ref{f5}(a). The robot arm with two degrees of freedom is adopted by fixing 2, 3, 5, 6, and 7 joints of the CRANE-X7 robot. The first and second links have masses and lengths of 0.651\text{kg}, 0.343\text{m}, and 0.543\text{kg}, 0.250\text{m}, respectively. The computer receives sensory information from the robot arm and force sensor, then computes control inputs for two actuators to regulate robot motions. This process is performed on Ubuntu 20.04.6 LTS. 

To demonstrate the feasibility and efficiency of the proposed method, we compare the proposed method with two methods based on admittance control. In fact, admittance control was widely applied in controlling the robot under pHRI tasks in \cite{c6.1, c7, c7.12, c7.13, c7.15} thanks to generating well-suited motions in compliance with interaction forces. Although the control parameters of the compared methods 1 and 2 are identical, as in Table \ref{table1}, the compared method 1 is integrated with the prediction model using the deep LSTM-based data-driven in \eqref{eq6.1}, \eqref{eq6.2}, and Fig. \ref{f3} with admittance control in \eqref{eq7}. Meanwhile, the compared method 2 is admittance control which solely utilizes human--robot interaction force as a trigger for movement generation as the following formulation ${{\mathbf{M}}_{d}}{{{\mathbf{\ddot{x}}}}_{a}}+{{\mathbf{D}}_{d}}{{{\mathbf{\dot{x}}}}_{a}}={{\mathbf{f}}_{ext}}$. The parameters for the proposed method and the compared methods 1, 2 are presented in Table \ref{table1}. Beyond the admittance parameters, the low-level control parameters of the proposed method are determined based on the conditions in Remark \ref{r2}, while these parameters of the compared methods 1 and 2 are fine-tuned to minimize tracking errors as in \cite{c7.12} through both simulations and experiments.

According to the aforementioned experimental setup in Fig. \ref{f5} and Table \ref{table1}, we conducted two stages, including preparation stage and human-in-the-loop experiment stage. The preparation stage was designed to collect co-carrying motion data and train the deep LSTM model for predicting human motion. Additionally, the experiment stage was implemented to demonstrate the effectiveness of the proposed method relative to the compared methods 1, 2.

\subsubsection{Preparation Stage} 
To train the deep LSTM model in Fig. \ref{f3} of the proposed method, we collected a dataset of the coupled motion between the human, robot partner, and co-transported object. During the data collection process, the robot partner under the proposed control method passively followed the movements of the human based on human forces, in which ${{{\mathbf{\ddot{\hat{x}}}}} }=\mathbf{0}$ and ${{{\mathbf{\dot{\hat{x}}}}}}=\mathbf{0}$ were set for \eqref{eq7}. 4 participants (4 males with an average age of 25.75 $\pm $ 2.06 SD) participated in the stage and they were instructed about the procedure. Firstly, each participant collaborated with the robot arm to grab a single object firmly at the initial position (O), as depicted in Fig. \ref{f5}(b).  Thereafter, the object was transported to one of three positions (A, B, C). This process was repeated 10 times for each position (A, B, C). In this stage, co-carrying trajectories for each trial were decided by participants without restriction. In this manner, we generated a dataset for the proposed method with 120 trials. The parameters of the deep LSTM model and training details are presented in Table \ref{table2}. After the training process, the loss value on the validation set reached $1.39 \times 10^{-4}$. Note that the compared method 1 also included the deep LSTM model in \eqref{eq6.1} and \eqref{eq6.2}. Therefore, we collected a separate dataset and trained the deep LSTM model for the compared method 1 by following the procedure and parameters in Table \ref{table2} as performed for the proposed method.

\begin{figure*}[!t]
\centering
\subfloat[]{{\includegraphics[width=3.5in]{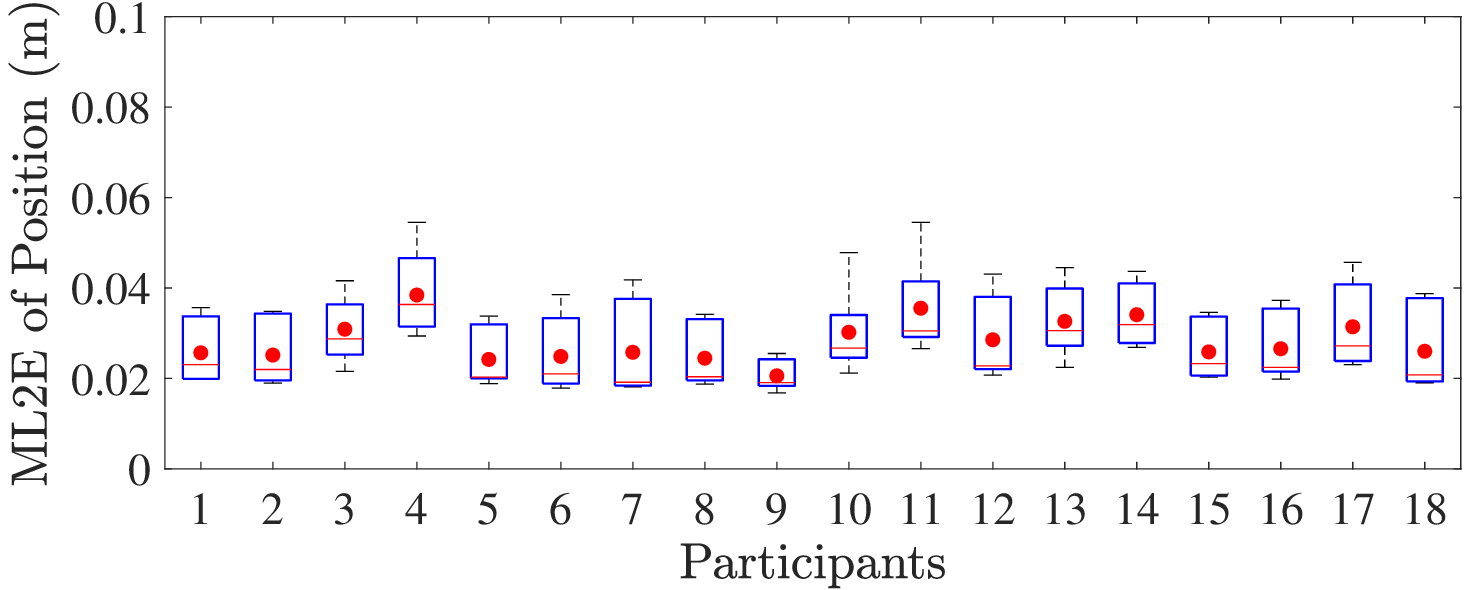}}%
\label{f6.1}}
\hfil
\subfloat[]{{\includegraphics[width=3.5in]{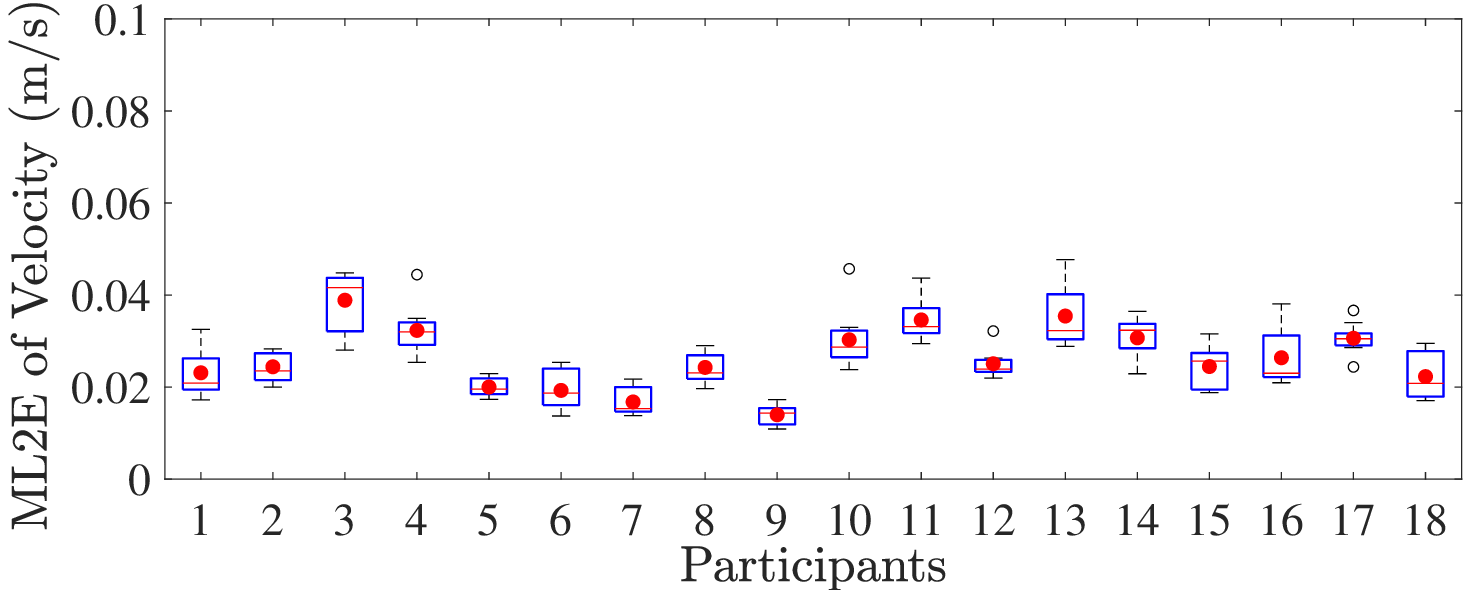}}%
\label{f6.2}}
\caption{Prediction performance of the deep LSTM model for each participant, (a) ML2E of position and (b) ML2E of velocity.}
\label{f6}
\end{figure*}

\subsubsection{Human-in-the-loop Experiment Stage} To validate the feasibility and efficiency of the proposed method, we compared it with the two baseline methods under the experimental setup in Fig. \ref{f5}(b), (c), and (d) via three co-carrying scenarios (O-B-A, O-B-C, O-B-D). Initially, the human collaborated with the robot partner at the starting position O, aiming to transport the object to position B, as illustrated in Fig. \ref{f5}(b). The object was then momentarily held at position B before being moved to one of the designated parking locations (A, C, or D), as shown in Fig. \ref{f5}(c). Finally, the object remained at the selected parking location (e.g., Fig. \ref{f5}(d)). For these scenarios, sudden changes decided by human intentions in achieving parking locations were considered. 18 participants (7 females and 11 males with an average age of 29.29 $\pm$ 4.93 SD) performed in the study and all provided informed consent. In the experiment, three trials were implemented for each scenario (O-B-A, O-B-C, or O-B-D) of each method (proposed method, compared method 1, and compared method 2) by each participant.  Therefore, each participant performed 27 trials. To mitigate potential learning and fatigue effects associated with the in-person approach, the order of the three methods and scenarios was randomized for each participant. After each method, the following questionnaire was administered using 7-point Likert scale to investigate the subjective evaluations from the participants:
\begin{enumerate}[label=(Q\arabic*)]
\item{Did you think the robot understands your motion intentions?}
\item{Did you feel the task lightly?}
\item{Did you think the task was easy to achieve parking locations?}
\end{enumerate}
Question Q1 aimed to determine whether the participants perceived that the robot accurately understood their motion intentions, reflecting the effectiveness of human–robot communication. Question Q2 explored how the task was experienced in terms of physical and cognitive effort, highlighting the impact of robotic assistance on workload. Question Q3 was designed to investigate participants’ perception of task difficulty in reaching the designated parking locations, providing insight into the robot’s support on task performance. The experiment was approved by the Ethics Review Committee of 
Nara Institute of Science and Technology (No. 2024-I-17).
\subsection{Analytical Method}
\begin{figure*}[!t]
\centering
\includegraphics[scale=0.40]{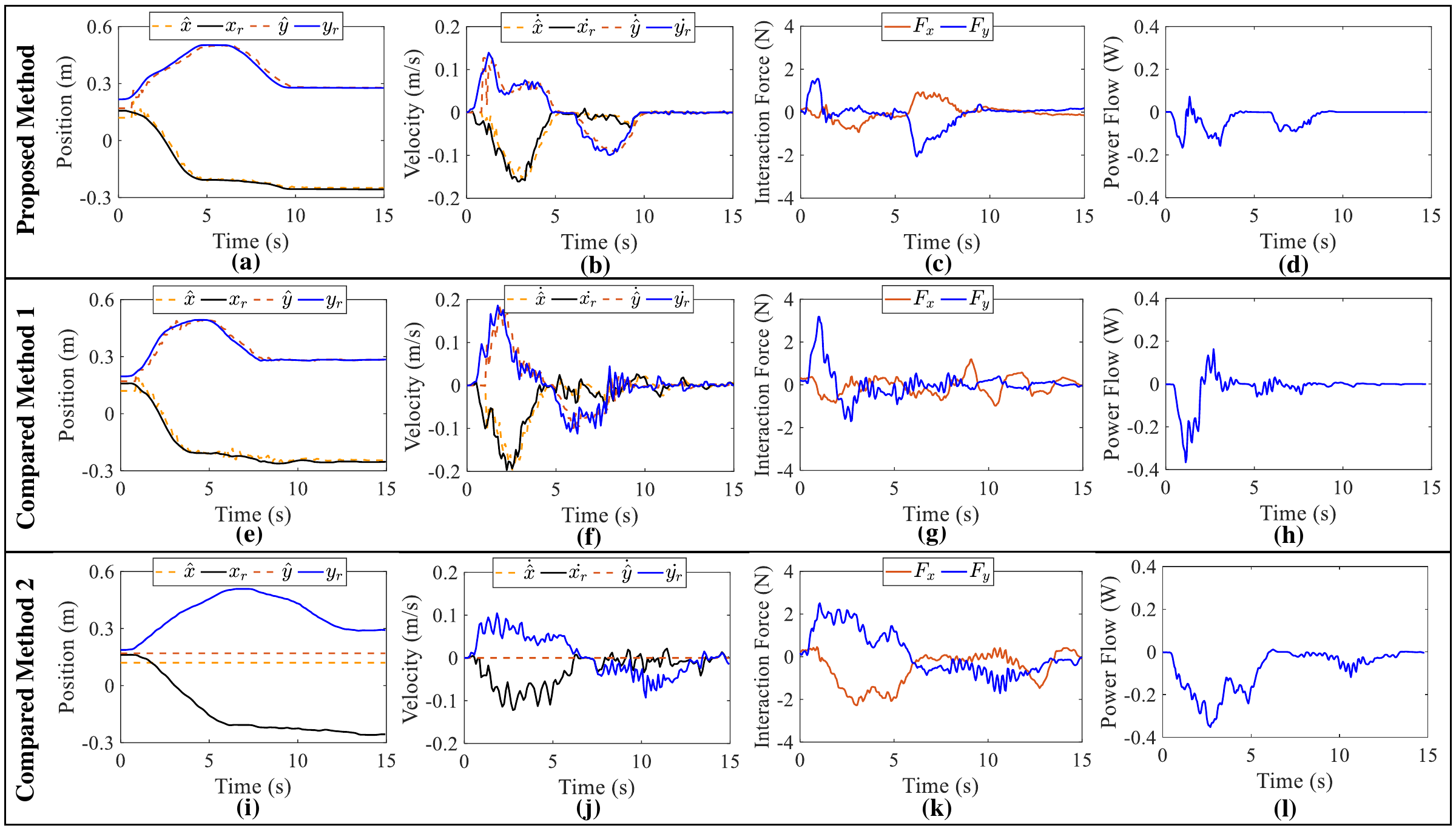}
\caption{Comparing three methods under Participant 9, including (a), (e), (i) are position and its prediction; (b), (f), (j) are velocity and its prediction; (c), (g), (k) are interaction force; (d), (h), (l) are power flow of the proposed method, the compared method 1, and the compared method 2, respectively.}
\label{f7}
\end{figure*}
In this section, we present analytical approaches to evaluate prediction performance and human--robot co-carrying efficiency for the proposed method in comparison with the two baseline methods. Firstly, prediction motion performance of the proposed method is evaluated by utilizing mean $L_2$-norm of error (ML2E) for position and velocity, including mean of ${{\left\| \left( {{{\hat{x}}}_{k}}-{{x}_{r,k}},{{{\hat{y}}}_{k}}-{{y}_{r,k}} \right) \right\|}_{2}}$ and mean of ${{\left\| \left( {{{\dot{\hat{x}}}}_{k}}-{{{\dot{x}}}_{r,k}},{{{\dot{\hat{y}}}}_{k}}-{{{\dot{y}}}_{r,k}} \right) \right\|}_{2}}$. In which ${{\hat{x}}_{k}},\,\,{{\hat{y}}_{k}},\,\,{{\dot{\hat{x}}}_{k}},$ ${{\dot{\hat{y}}}_{k}}$ and ${{x}_{r,k}},\,{{y}_{r,k}},\,{{{\dot{x}}}_{r,k}},$ ${{{\dot{y}}}_{r,k}}$ are predicted and actual positions and velocities on x and y-axes, respectively. Additionally, the data of interaction force and power flow are analyzed in magnitude and variation viewpoints. For magnitude, discrete-time integration of resultant force (DI-RF), discrete-time integration of absolute power flow (DI-APF), root mean square of resultant force (RMS-RF), and root mean square of power flow (RMS-PF) are used. Furthermore, we investigate the variation feature based on the range of resultant force and power flow $\left(\text{Range-RF}, \text{Range-PF}\right)$ calculated by the difference between the maximum and minimum values as well as standard deviation of resultant force (SD-RF) and power flow (SD-PF).

Moreover, statistical tests are implemented to compare the proposed method with two methods based on representative data, which are the mean of each above metric for each participant. Examination via the Shapiro–Wilk test indicated that the data did not satisfy a normal distribution; therefore, the Friedman test and the Wilcoxon signed-rank test with Holm correction are utilized in our paper. The data analysis procedure was conducted in two steps. Initially, the main effect of the data was examined by the Friedman test. If the main effect was significant, the post-hoc Wilcoxon signed-rank tests with Holm correction were performed to compare the interaction force and power flow of all pairs of the three methods. In addition, statistical tests for subjective results are also conducted by following the aforementioned process.

\subsection{Experiment Results}


The predicted position and velocity performances of each participant were computed by utilizing ML2E and depicted in Fig. \ref{f6}. According to the data obtained from conducting three trials for each scenario, the distribution of ML2E for predicted position and velocity was observed through boxplots in Figs. \ref{f6.1} and \ref{f6.2}, respectively. The variation in ML2E indicated that the predictions of the deep LSTM model were influenced by both the human intention during each trial and the individual characteristics of each participant. Based on the metrics from 18 participants, the mean ML2E for predicted position and velocity were varied within the intervals $\left[ 0.014,0.039 \right]\,\text{m}$ and $\left[ 0.021,0.038 \right]\, \text{m/s}$, respectively. The presence of predicted errors could be attributed to the variability in human intentions and the inherent measurement noise, which cannot be completely eliminated in real-world applications. In this context, the requirement for robotic controllers is to ensure safe interaction and co-carrying task completion while reducing human efforts.

\begin{figure*}[!t]
\centering
\includegraphics[scale=0.4]{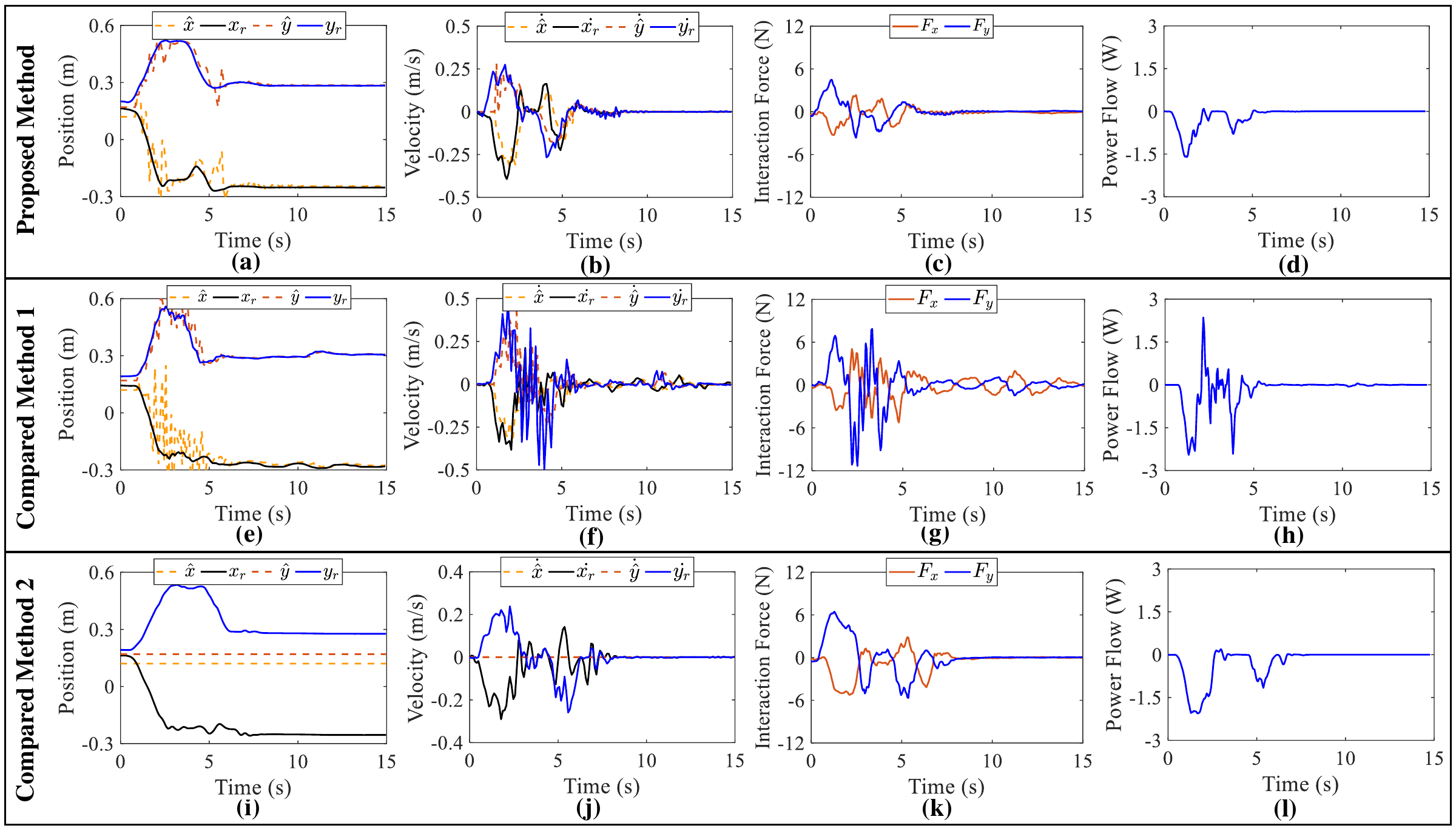}
\caption{Comparing three methods under Participant 3, including (a), (e), (i) are position and its prediction; (b), (f), (j) are velocity and its prediction; (c), (g), (k) are interaction force; (d), (h), (l) are power flow of the proposed method, the compared method 1, and the compared method 2, respectively.}
\label{f8}
\end{figure*}
\begin{figure}[!t]
\centering
\includegraphics[scale=0.35]{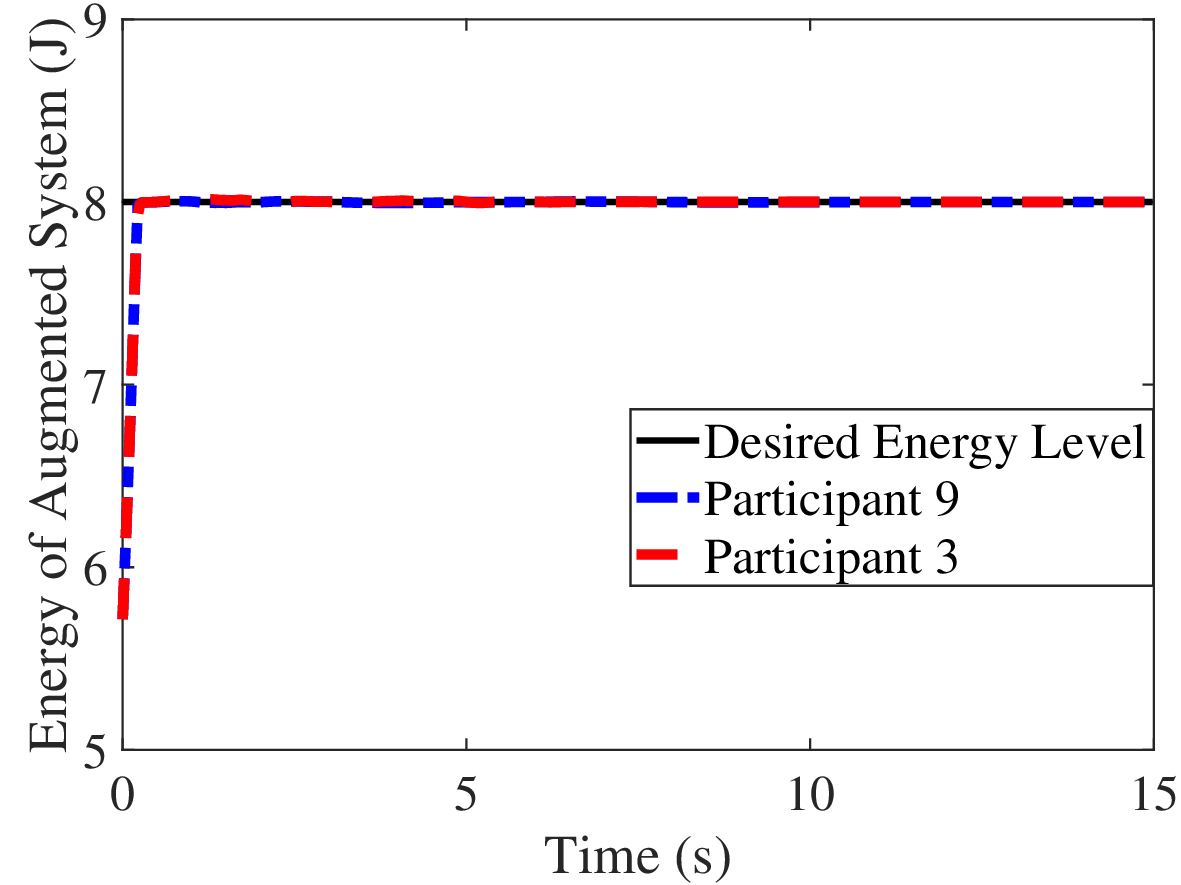}
\caption{Energy level convergence in the augmented system under collaboration with Participant 9 and 3 for the proposed method.}
\label{f78}
\end{figure}
To specifically observe the responses in the task, we first present the results of two representative participants in achieving the same parking location (O-B-D). According to the prediction performance in Fig. \ref{f6}, two participants with the lowest and highest mean ML2E were selected, corresponding to participant 9 and participant 3, respectively. For participant 9, the deep LSTM model of the proposed method and the compared method 1 demonstrated a relatively high-level of accuracy in predicting position and velocity, as shown in Figs. \ref{f7}(a), (b), (e), (f). Meanwhile, in Figs. \ref{f7}(i) and (j), the motion prediction was not performed because the compared method 2 was admittance model without prediction model. In this situation, the amount of interaction force and power flow of the proposed method and the compared method 1 was lower than that of the compared method 2, as illustrated in Figs. \ref{f7}(c), (g), (k), and (d), (h), (l). In addition, participant 3 also performed the co-carrying task with higher velocity in comparison with participant 9's data in Fig. \ref{f7}; therefore, the three methods in Fig. \ref{f8} displayed the higher interaction force and power flow.  Based on the results of participant 3 in Fig. \ref{f8}, the proposed method was evaluated promising solution for the co-carrying task with the smallest interaction force and power flow. Moreover, the energy levels in the augmented system when collaborating with participants 9 and 3 were also exhibited in Fig. \ref{f78}, which converged to the desired energy level $\left(k_{a}^{d}=8\,\text{J}\right)$ after approximately 0.25s. Note that the feature was also observed in collaborating with the other participants. However, limiting the analysis to the observation of special cases is inadequate for drawing comprehensive conclusions about the proposed controller. Thus, analytical approaches and statistical tests, which are given in Section IV. B, needs to be considered in the next part. 

\begin{figure}[!t]
\centering
\includegraphics[scale=0.4]{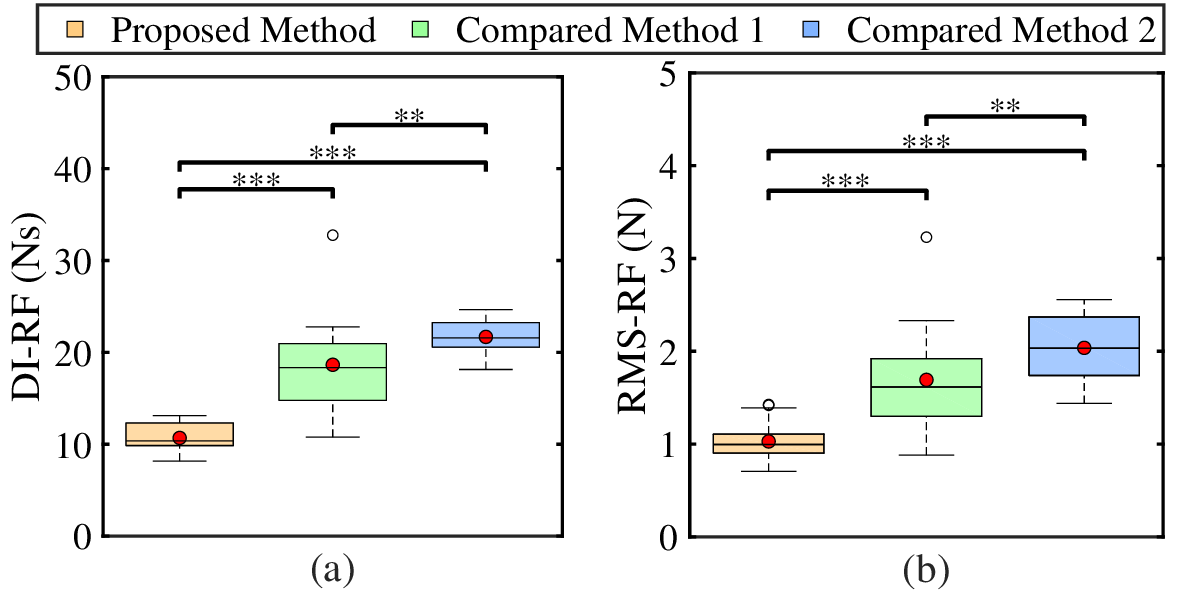}
\caption{Comparing DI and RMS of resultant interaction force $\left( \text{**}:\text{p}<0.01,\text{***}:\text{p}<0.001 \right)$.}
\label{f9}
\end{figure}

\begin{figure}[!t]
\centering
\includegraphics[scale=0.4]{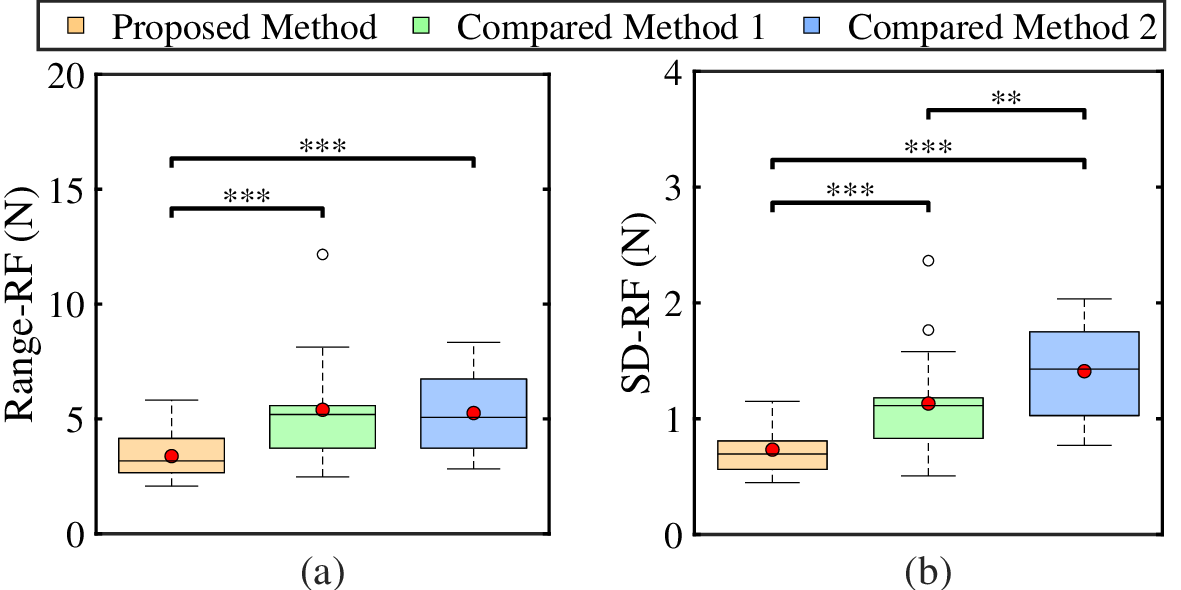}
\caption{Comparing range and SD of resultant interaction force $\left( \text{**}:\text{p}<0.01,\text{***}:\text{p}<0.001 \right)$.}
\label{f10}
\end{figure}

\begin{figure}[!t]
\centering
\includegraphics[scale=0.4]{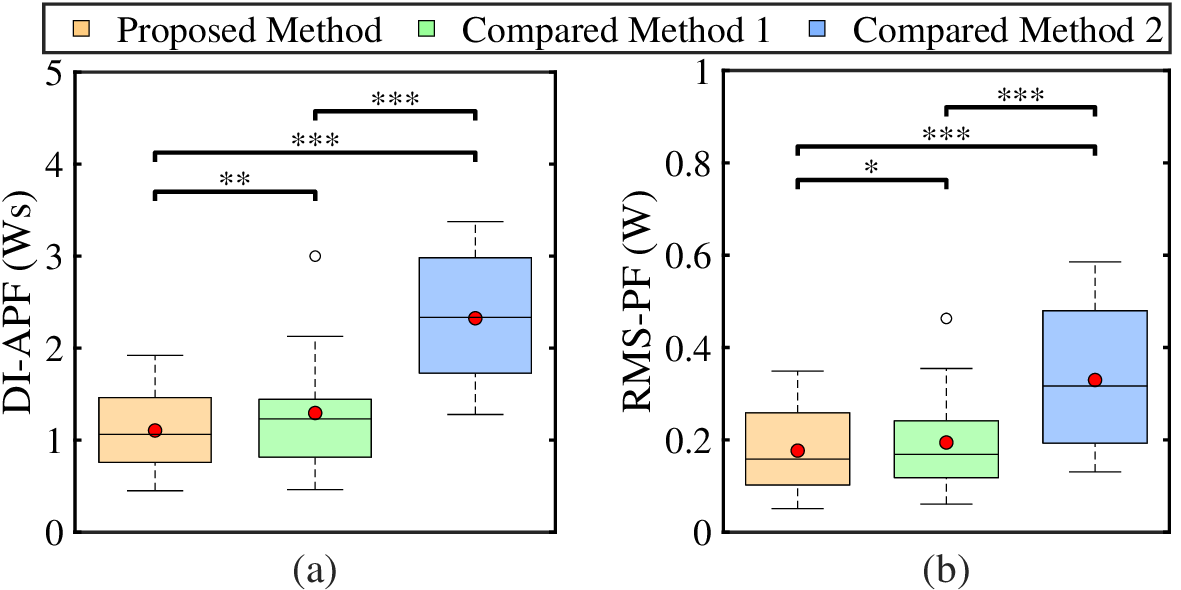}
\caption{Comparing DI of absolute power flow and RMS of power flow $\left( \text{*}:\text{p}<0.05,\text{**}:\text{p}<0.01,\text{***}:\text{p}<0.001 \right)$.}
\label{f11}
\end{figure}

\begin{figure}[!t]
\centering
\includegraphics[scale=0.5,trim=36 0 35 30,clip]{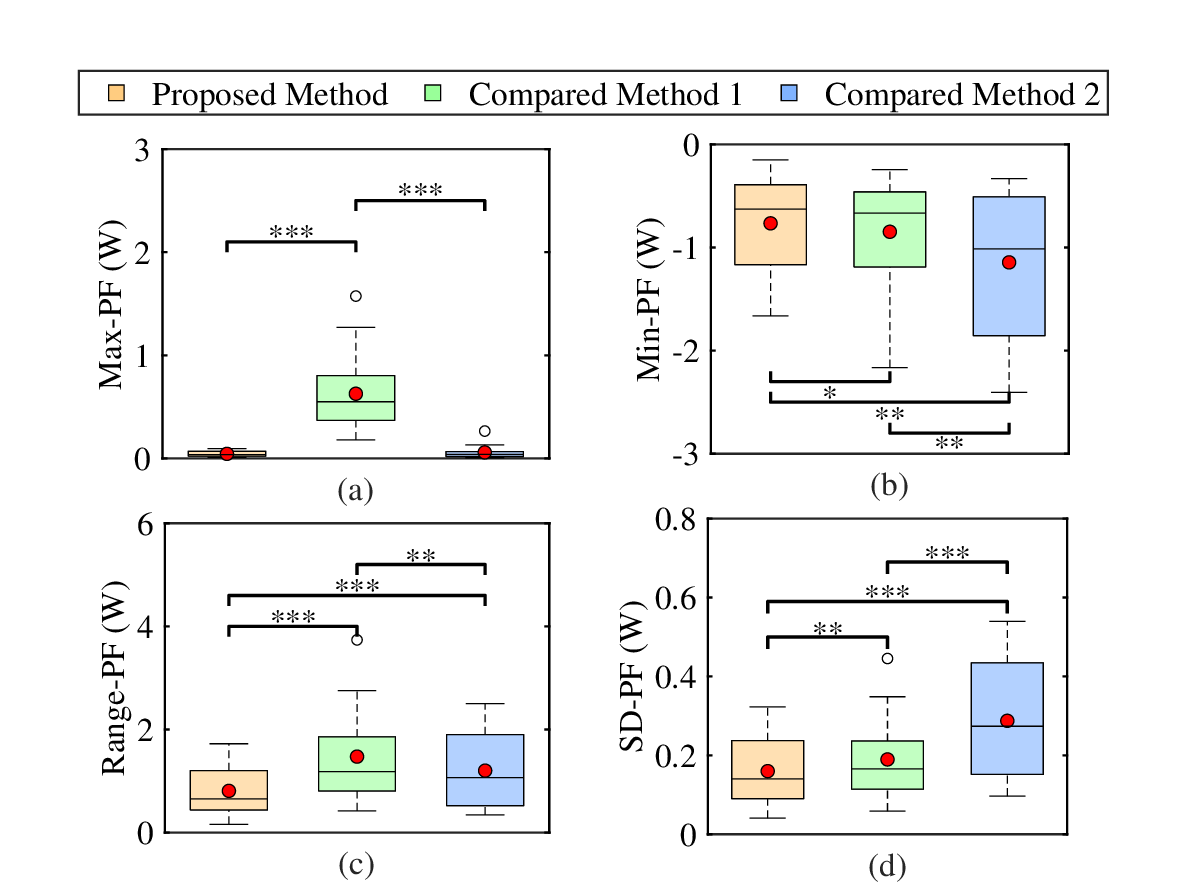}
\caption{Comparing maximum, minimum, range, and SD of power flow $\left( \text{*}:\text{p}<0.05,\text{**}:\text{p}<0.01,\text{***}:\text{p}<0.001 \right)$.}
\label{f12}
\end{figure}

According to the results in Table \ref{table3}, the Friedman tests of each metric in Section IV. B revealed statistically exceptional strong differences between the three control methods $\left( \text{***}:\text{p}<0.001 \right)$. Therefore, post-hoc Wilcoxon signed-rank tests with Holm correction were implemented to compare all pairs of the three methods.  For the results in Figs. \ref{f9} and \ref{f10} related to the compared results of magnitude and variation of interaction forces, the proposed method showed exceptionally smaller DI-RF, RMS-RF, Range-RF and SD-RF metrics in comparison with the compared method 1 $\left( \text{z}=-3.72,\,\text{p}=0.00059, \text{for all metrics} \right)$ and the compared method 2 $\left( \text{z}=-3.72,\,\text{p}=0.00039, \text{for all metrics} \right)$. The metrics of DI-RF in Fig. 9(a), RMS-RF in Fig. 9(b), and SD-RF in Fig. 10(b) for the compared method 2, were higher than those for the compared method 1, with corresponding statistical results of $\left( \text{z}=2.63,\,\text{p}=0.0084 \right)$, $\left( \text{z}=3.03,\,\text{p}=0.0025 \right)$, and $\left( \text{z}=3.16,\,\text{p}=0.0016 \right)$. Meanwhile, no significant differences were observed between the compared methods 1 and 2 about Range-RF $\left( \text{z}=-0.29,\,\text{p}=0.85 \right)$ in Fig. 10(a).
In addition to comparisons of interaction force,  the power flow within the human and robot revealed the following results: As depicted in Fig. \ref{f11}, the DI-PF and RMS-PF metrics of the proposed method and the compared method 1 were significantly smaller than those for the compared method 2, with $\left( \text{z}=-3.72,\,\text{p}=0.00059, \text{for all metrics} \right)$ and $\left( \text{z}=-3.72,\,\text{p}=0.00039, \text{for all metrics} \right)$, respectively. The proposed method also exhibited smaller values than the compared method 1 in terms of DI-PF $\left( \text{z}=-2.72,\,\text{p}=0.0065 \right)$  and RMS-PF $\left( \text{z}=-2.46,\,\text{p}=0.014 \right)$. In Fig. \ref{f12}a, the maximum power flow for the compared method 1 was bigger than the proposed method $\left( \text{z}=-3.72,\,\text{p}=0.00059 \right)$ and the compared method 1 $\left( \text{z}=-3.72,\,\text{p}=0.00039 \right)$, while no significant differences were given between them $\left( \text{z}=-0.98,\,\text{p}=0.33 \right)$.
\begin{table*}[t]
\centering
\caption{Friedman test results for each analytical method in Section IV. B reflected the main effects of three methods}
\label{tab:friedman_test}
\begin{tabular}{|c|c|c|c|c|c|c|c|c|c|c|}
\hline
{\parbox[c][0.6cm][c]{1.8cm}{\centering \textbf{ {} \\
Friedman Test\\Results}}} & \multicolumn{4}{c|}{\textbf{Interaction Force}} & \multicolumn{6}{c|}{\textbf{Power Flow}} \\ \cline{2-11} \rule{0pt}{3ex}
 & \text{DI-RF} & \text{RMS-RF} & \text{Range-RF} & \text{SD-RF} & \text{DI-APF} & \text{RMS-PF} & \text{Max-PF} & \text{Min-PF} & \text{Range-PF} & \text{SD-PF} \\ \hline
\rule{0pt}{3ex}  
${{\chi }^{2}}\left( 2 \right)$ & 29.78 & 32.44 & 27.11 & 32.44 & 31.00 & 29.78 &  27.11 & 21.78 & 25.33 & 32.44 \\ \hline
\rule{0pt}{3ex}  
$\text{p}$ & $3.14\mathrm{e}^{-7}$ & $9.01\mathrm{e}^{-8}$ & $1.30\mathrm{e}^{-6}$ & $9.01\mathrm{e}^{-8}$ & $1.86\mathrm{e}^{-7}$ & $3.42\mathrm{e}^{-7}$ & $1.30\mathrm{e}^{-6}$ & $1.87\mathrm{e}^{-5}$ & $3.16\mathrm{e}^{-6}$ & $9.01\mathrm{e}^{-8}$ \\ \hline
\end{tabular}
\label{table3}
\end{table*}

The minimum power flow of the proposed method was closer to zero compared to those of the compared methods 1 $\left( \text{z}=2.07,\,\text{p}=0.039 \right)$ and the compared method 2 $\left( \text{z}=3.59,\,\text{p}=0.0010 \right)$, as well as the metric for the compared method 2 exhibited a larger deviation from zero than that of the compared method 1 $\left( \text{z}=-2.85,\,\text{p}=0.0087 \right)$, as illustrated in Fig. \ref{f12}(b). Regarding Figs. \ref{f12}(c) and \ref{f12}(d), the proposed method revealed exceptionally smaller Range-PF and SD-PF values compared to the compared method 1 $\left(\text{z}=-3.72,\,\text{p}=0.00059\right)$, $\left(\text{z}=-3.11,\,\text{p}=0.0018\right)$ and the compared method 2 $\left(\text{z}=-3.51,\,\text{p}=0.00091\right)$, $\left(\text{z}=-3.72,\,\text{p}=0.00059\right)$. While the Range-PF of the compared method 1 was significantly higher than the compared method 2 $\left( \text{z}=2.59,\,\text{p}=0.0096 \right)$, the opposite trend was observed in the case of SD-PF metric $\left( \text{z}=-3.72,\,\text{p}=0.00039 \right)$. 

\begin{figure}[!t]
\centering
\includegraphics[scale=0.45,trim=30 0 50 5,clip]{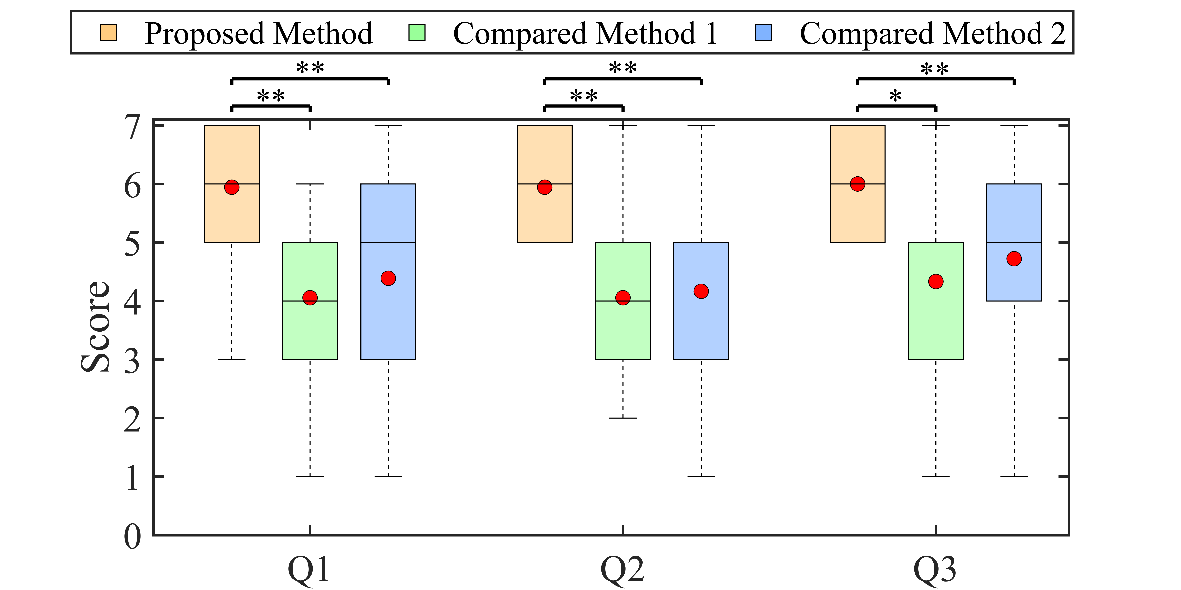}
\caption{The scores of the questions Q1, Q2, and Q3 $\left( \text{*}:\text{p}<0.05,\text{**}:\text{p}<0.01 \right)$.}
\label{f13}
\end{figure}

Other Friedman tests of subjective results from three questions were conducted and revealed the statistically significant main effect of the three methods (Q1: $\chi \left( 2 \right)=20.36,p<0.001$, Q2: $\chi \left( 2 \right)=20.22,p<0.001$, and Q3: $\chi \left( 2 \right)=13.41,p<0.01$). Subsequently, a series of Wilcoxon signed-rank tests with Holm correction was applied to compare all pairs of the three methods under each question. Specifically, the proposed method exibited significantly higher scores compared with the compared method 1 (Q1: z = 3.45, p = 0.0017; Q2: z = 3.09, p = 0.0041; Q3: z = 2.71, p = 0.013) and compared method 2 (Q1: z = 3.35, p = 0.0016; Q2: z = 3.55, p = 0.0011; Q3: z = 2.97, p = 0.0087), as illustrated in Fig. \ref{f13}. Meanwhile, no significant differences were found in the remaining comparisons in Fig. \ref{f13} (Q1: z = -0.66, p = 0.51; Q2: z = -0.23, p = 0.82; Q3: z =-0.72, p = 0.47). The consistently higher scores across subjective results suggest that the proposed method enhances human--robot communication, reduces workload, and improves task performance in reaching the designated parking locations compared with the other methods.

\section{Discussion}


In the present study, we constructed a cooperation control framework using the reference generator and low-level controller for human--robot co-carrying tasks as well as examined its effectiveness through both theoretical and experimental viewpoints.

The reference generator predicted human motion during the task and adjusted the predicted motion in response to interaction forces in the presence of conflicts. In this manner, when the human motion was accurately predicted, the robot could proactively assist in the task, while simultaneously reducing conflicts due to prediction errors. This finding was possible to achieve proactive robotic behavior without the need for discrete switching between control modes \cite{c7,c7.13}. Notably, although belonging to the admittance control family, A. Sharkawy et al. \cite{c7.12} and E. Shahriari et al. \cite{c7.6} primarily concentrated on adjusting admittance parameters rather than integrating human motion prediction, as in the present study. These approaches demonstrated the effectiveness in pHRI tasks. While variable admittance control in \cite{c7.12,c7.6} provided flexibility in responding to human guidance, the predicted motion-based admittance control in this paper enabled proactive coordination. Therefore, combining the adjusted admittance parameters mechanism and predicted human motion will be an important focus of future research. 

For the low-level controller, an improved time-varying PVFC was proposed to minimize motion errors, maintain energy levels, and guarantee semi-passivity in a controlled manner of the closed-loop robotic system, thereby enhancing task performance and safety interaction simultaneously. Meanwhile, E. Dincer et al. \cite{c7} ignored the low-level control to mainly focus on admittance model. M. Ma et al. \cite{c6.1} and D. Sirintuna et al. \cite{c7.15} constructed tracking control to only minimize motion errors without energy consideration, as the proposed method, which may lead to unsafe situations for humans in the presence of conflicts. Unlike the semi-pasive property in the present study, enforcing passivity at all times may result in conservative behaviors of the robot that, similar to previous approaches \cite{c18,c18.11}, limit the effectiveness of robotic assistance. These comparisons showed that the proposed method achieved a balance between stability, safety, and task efficiency.

In addition, human-in-the-loop experiments in Section IV further demonstrated the effectiveness of the proposed framework. Through comparative analysis using the metrics Section IV.B.2 and subjective results, the proposed framework revealed superior effectiveness compared to the compared methods 1 and 2. For the objective evaluations, in Figs. \ref{f9} and \ref{f11}, the proposed method exhibited the lowest magnitude of the resultant interaction force and power flow compared to the compared methods 1 and 2, thereby reducing human workload and enhancing the safety level of human--robot interaction. Furthermore, the variation of the resultant interaction force and power flow can also be observed in Figs. \ref{f10} and \ref{f12} through the metrics of Range and SD. The proposed method facilitated a smoother and more effortless execution of co-carrying tasks compared to the other methods, owing to its minimal variation. For the subjective results, the proposed method achieved higher scores than compared methods 1 and 2 across all three subjective measures related to human--robot communication, workload reduction, and task performance, as illustrated in Fig. \ref{f13}. Overall, these findings highlighted that the advantages of the proposed method arose from both the reference generator and the low-level controller, reinforcing theoretical results.

As presented in Section IV, the effectiveness of the proposed framework was evaluated through two-dimensional human-in-the-loop experiments on the XY plane. While three-dimensional experiments were not performed, this section also discusses the potential extension to three-dimensional scenarios, highlighting the corresponding modifications required in the deep LSTM model and the improved time-varying PVFC. First, incorporating additional translational (z-axis) and rotational components increases the input dimensionality for the deep LSTM model. It is necessary to consider increasing the size of the training dataset, the number of hidden units per LSTM layer, and the number of stacked LSTM layers in order to effectively capture complex spatiotemporal correlations. Second, although the improved time-varying PVFC presented in Section III.B was designed for a general $n$-link robotic model, defining the corresponding dynamic model in the form of \eqref{eq4} and the associated Jacobian matrix remains necessary to account for the additional z-axis motion. It should be noted that the gravity-compensation mechanism must be incorporated alongside the proposed method in three-dimensional experiments, similar to the approach \cite{c18.11}. Thus, by refining the aforementioned aspects, the proposed framework can be extended to three-dimensional scenarios to assess its adaptability in more complex tasks in future work.

In addition, the present study had some limitations. As discussed above, the human-in-the-loop experiments in the present study were conducted in two-dimensional scenarios constrained to the OXY plane. It is thus necessary to verify the adaptability of the proposed method and draw generalized conclusions across a broader set of three-dimensional spatial motion scenarios. Moreover, the training dataset of the deep LSTM model and the participant pool in the main experiments exhibited an imbalance in age groups. To enhance the generalizability of the findings, future work should involve validation of the proposed method with datasets that encompass a wider variety of participants. Furthermore, the present study focused on the human--robot co-carrying task for a small transported object on a horizontal surface, thereby ignoring the gravity effect, weight, and size of the object. However, these characteristics could influence the feasibility and efficiency of the proposed framework when considering large or heavy objects in the co-carrying task that cannot be managed solely by either the human or the robot partner. Therefore, further validation is needed to examine the applicability of the proposed method across different object sizes and weights for future research. 

\section{CONCLUSION AND FUTURE WORK}
In this paper, a continuous cooperative control framework incorporating a reference generator and low-level control was proposed and applied for human--robot co-carrying tasks. The reference generator not only predicted human intention but also re-planned it in the event of prediction errors. An improved PVFC was developed to guarantee the passivity of the closed-loop system in a controlled manner, maintain energy level, regulate power flow, and track the co-carrying path. The present method achieved a balance between proactive assistance and passive safety in human--robot collaborative object transportation tasks. Human-in-the-loop experiments were conducted and demonstrated the effectiveness of the proposed method through both objective and subjective results.

In the future, the proposed cooperation control method needs to be extended to be applicable to other human--robot cooperation systems such as handover, assembly, teleoperation, sawing, and so on. In addition, the proposed framework in the present study will be enhanced to address real-world human--robot co-carrying tasks, such as the cooperative transportation of heavy or large objects.

\section*{Acknowledgments}
This work was partially supported by JSPS KAKENHI (Grant Number 24H00298), Japan.



%

\bibliographystyle{IEEEtran}
\bibliography{mybibtex}

\end{document}